\documentclass[runningheads]{llncs}
\usepackage[T1]{fontenc}
\usepackage{graphicx}
\usepackage{booktabs}
\usepackage[misc]{ifsym}
\newcommand{\corr}{(\Letter)}
\usepackage{amsmath}
\usepackage{amssymb}
\usepackage{hyperref}
\usepackage{cleveref}
\usepackage{url}
\usepackage{makecell}
\usepackage{subcaption}
\usepackage{xcolor}

\usepackage{wrapfig}
\usepackage{multirow}
\usepackage{multicol}
\usepackage{titletoc}
\usepackage{tcolorbox}

\let\citep\cite

\crefname{figure}{Fig.}{Figs.}
\crefname{table}{Tab.}{Tabs.}
\crefname{section}{Sec.}{Secs.}

\begin{document}

\title{Zero-Shot Test-Time Canonicalization\\using Out-of-Distribution Scoring}

\titlerunning{Zero-Shot Test-Time Canonicalization}


\author{Dominik Lindner\inst{1} \corr \and
Johann Schmidt\inst{1} \and
Tom Siegl\inst{2} \and Martin Becker\inst{3} \and Sebastian Stober\inst{1}}

\authorrunning{D. Lindner et al.}

\institute{Artificial Intelligence Lab, Otto-von-Guericke University Magdeburg, Germany \email{\{dominik.lindner, johann.schmidt, stober\}@ovgu.de}
\and
University of Rostock, Germany \email{tom.siegl@uni-rostock.de}
\and
Becker Lab, University of Marburg, Germany \email{martin.becker@uni-marburg.de}
}

\tocauthor{Dominik Lindner,Johann Schmidt, Tom Siegl,Martin Becker, Sebastian Stober}
\toctitle{Zero-Shot Test-Time Canonicalization using Out-of-Distribution Scoring}

\maketitle              

\begin{abstract}
Pretrained vision models often misclassify inputs that are rotated, scaled, or sheared, even though these affine transformations leave the object class unchanged. 
Robustness is usually restored either by building equivariance into the architecture or by retraining with augmentation, both of which require changing or retraining the model. 
Test-time canonicalization instead leaves the classifier untouched.
It undoes the transformation of each input, mapping it to a canonical form near the training distribution before classification. 
Existing canonicalizers, however, rely on a narrow set of logit-based energy scores and bespoke search procedures, leaving the design space of scoring functions and optimizers unexplored. 
We reframe canonicalization as out-of-distribution (OOD) detection, which lets any OOD score serve as the energy minimized over transformations. 
Across benchmarks ranging from handwritten characters and sketches to natural images and 3D point clouds, we systematically evaluate around twenty OOD scores and nine search algorithms, finding that distance-based scores paired with random search and local refinement perform best overall. 
Because canonicalizing an already-aligned input can hurt accuracy, we add a gated mechanism that transforms an input only when its OOD score indicates this is needed, preserving most in-distribution accuracy while retaining the robustness gains on transformed inputs.
Code is available at \url{github.com/johschm/its}.

\end{abstract}

\begin{figure}[th]
    \centering
    \includegraphics[width=1.0\linewidth]{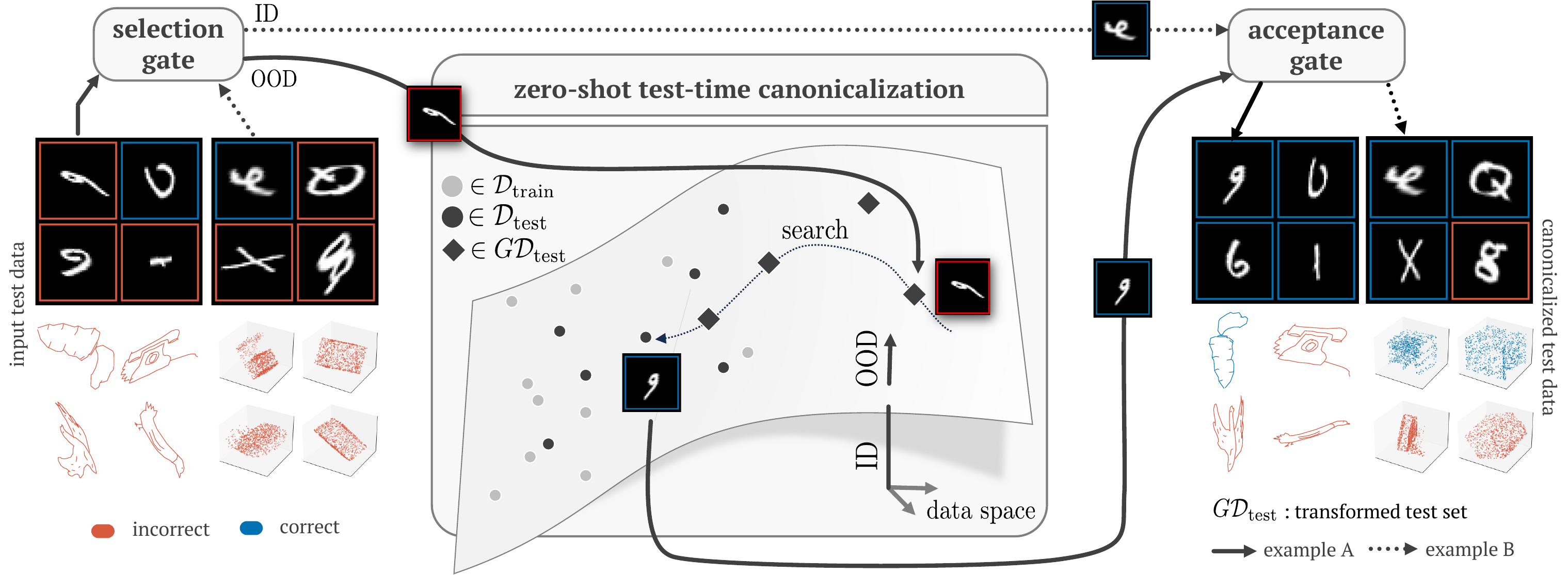}
    \caption{
    During test-time in-distribution (ID) samples are passed directly to the downstream model, while out-of-distribution (OOD) samples are passed to a zero-shot canonicalization module. 
    This module uses a search heuristic to minimize the OOD score of the sample by applying affine transformations.
    Only if this decreases the OOD score the result is passed to the downstream model.}
    \label{fig:pitch}
    \vspace{-0.3cm}
\end{figure}

\section{Introduction}
For a model to be reliable, its predictions should remain stable under class-preserving transformations of the input.
A rotated, rescaled, or sheared image of an object still depicts the same object.
Such transformations are common at deployment time, for instance in microscopy, aerial imagery, or sketch and document analysis, where the upright pose seen during training cannot be assumed.
Yet many vision models are not robust to them~\citep{Engstorm2019,SIScore}, and retraining a deployed model for every such setting is often impractical.
A strong form of robustness is invariance, where the predicted class remains unchanged under input transformations \cite{Jang2023}.
The most widely used method to increase robustness without altering the model architecture is \emph{data augmentation}, which adds semantically redundant, transformed samples to the training set to encourage learning robust filters \cite{Huang2021}.
Invariance can be guaranteed by equivariant models \cite{SteerableCNNs} using group-pooling \cite{GCNN}, which constrain the architecture through explicit inductive biases.
This, unfortunately, is very constraining, as such an inductive bias has to be designed for the specific transformation and the model has to be trained from scratch.
The downstream prediction can be made invariant without such constraints by averaging over all transformations \citep{ICLR2025_b36dc39b}.
Subsequent works reduce this to an average over a subspaces \citep{Puny2022}, but the computational cost remains prohibitive.
This is closely related to Test-time Augmentation (TTA), where predictions are also averaged over transformed inputs.
However, they assume a heuristic distribution over transformations and do not guarantee invariance \citep{Kim2026}.

An alternative approach to spatial robustness is \emph{canonicalization}, which seeks to transform inputs into a canonical form by ``undoing'' group transformations prior to inference. 
Crucially, canonicalization does not constrain the downstream architecture, which can therefore be treated as fixed.
Early canonicalizers learned to regress transformation parameters directly from data \cite{STN}.
However, this can lead to fragile predictions even with modern backbones \cite{ConstrSTN}.
Later approaches formalized canonicalization as the selection of a canonical representation for an input sample by minimizing an equivariant scoring function over group transformations \cite{Kaba2023,Mondal2023}.
These methods typically rely on equivariant features or architectures for the canonicalizer ~\cite{GCNN} to ensure consistent scoring across transformed inputs.
More recent work relaxes the requirement of equivariant canonicalizers by introducing training procedures for non-equivariant networks \cite{Panigrahi2024}, but such approaches still require substantial training effort.
Many learning-based canonicalizers further assume or benefit from training data that exhibits consistent alignment or a canonical pose distribution \cite{Mondal2023}.
This assumption is often violated in real-world datasets, where object pose may vary significantly. 
Recent work addresses this by progressively re-aligning the training data during learning \cite{Schmidt2025}, but at the cost of additional training complexity. 
This raises a practical question: can canonicalization be obtained without training at all?
Recent methods therefore formulate canonicalization as a test-time optimization problem, where a transformation is selected by minimizing a score defined directly on the downstream classifier's outputs \citep{ITS}. 
However, existing canonicalization methods are limited by their narrow choice of energy functions, reliance on bespoke and costly search procedures, and a tendency to degrade in-distribution (ID) accuracy when inputs are already aligned.
Existing approaches employ logit-based energies \citep{ITS} or heuristic criteria \citep{Shumaylov2025} that favor transformations producing confident ID predictions. 

We propose the hypothesis that OOD scores constitute suitable energy functions for canonicalization.
The underlying assumption is that a canonical form lies in a high-density region of the training distribution and therefore attains a low OOD score, while transformed inputs move off this manifold and score higher.
Minimizing an OOD score over the transformation group should thus recover a canonical form.
This is a necessary rather than a sufficient condition.
A low score favors transformations that appear typical, which need not coincide with the correct semantic pose, especially for classes with inherent pose ambiguity.
We therefore treat it as an empirical hypothesis and test it across a broad range of scores and datasets.
If it holds, the extensive body of OOD detection literature can be leveraged directly, broadening the design space for canonicalization.
Moreover, the optimization strategy used to search over transformations is equally important.
Prior work relies on bespoke search procedures tailored to specific scoring functions \citep{ITS,Singhal2025}, including Lie-algebra-based methods for continuous transformation groups \citep{Shumaylov2025,Kim2026}. 
While effective, these approaches introduce additional complexity.
Based on prior work, it remains unclear whether simpler global optimization strategies, such as simulated annealing or random search, which have shown benefits in other domains \cite{Schmidt2024b}, can achieve competitive performance for common transformations such as affine transformations.
Another issue of current canonicalization frameworks is the degradation of ID accuracy when inputs are already aligned \citep{ITS}, as incorrect or unnecessary transformations can distort features. 
This presents a practical challenge, since most real-world inputs are typically close to the training distribution and only a minority exhibit large transformations.
\newpage
Our main contributions to these problems are:
\begin{enumerate}
    \item We systematically benchmark OOD scores as energy functions for canonicalization, addressing the limited exploration of energy functions in prior work.
    We find that the best-performing score varied across datasets.
    \item We show that prior search procedures can be replaced by simpler global optimizers (e.g., random search, simulated annealing) without sacrificing performance, with random search with local refinements working best overall.
    \item We introduce a lightweight OOD-based gating mechanism that skips unnecessary canonicalization, resolving the accuracy trade-off in current frameworks by preserving performance on both ID and OOD inputs.
\end{enumerate}

\section{Preliminaries}
Let $(x, y) \sim \mathcal{D}$ denote input-label pairs, where $x \in \mathcal{X}$ and $y \in \mathcal{Y}$. 
Many vision tasks admit class-preserving input transformations: 
a rotated image of a fish remains a fish.
To study such transformations formally, we use group theory.
The following introduces the key concepts needed (see \Cref{app:preliminaries} and \cite{Bronstein2021,Hall2015} for more details).
A \emph{group} $G$ is a set equipped with an associative binary operation, an identity element, and inverses for every element. 
We restrict to matrix Lie groups \cite{Hall2015}, so every $g \in G$ is a matrix, and the group operation is matrix multiplication. 
A \emph{subgroup} $H \leq G$ is a subset that is itself a group under the inherited operation.
A \emph{group action} of $G$ on $\mathcal{X}$ is a map $(g, x) \mapsto g \cdot x$.
A prominent example is the affine group $\mathrm{Aff}_2(\mathbb{R})$, whose elements (any combination of rotation, scaling, shearing, and translation) act on pixel coordinates in homogeneous coordinates via $3\times 3$ matrix multiplication.
The \emph{orbit} of $x$ under $G$ is
$Gx = \{\, g \cdot x \mid g \in G \,\}$, which collects all transformed versions of $x$. 
The \emph{stabilizer} of $x$ is
$G_x = \{\, g \in G \mid g \cdot x = x \,\}$,
the subgroup of transformations that leave $x$ fixed. 
For $H \leq G$ and $g \in G$, the \emph{coset} $gH = \{gh \mid h \in H\}$ partitions $G$ into equivalence classes.
A function $f \colon \mathcal{X} \to \mathcal{Y}$ is $G$-equivariant if $f(g \cdot x) = g \cdot f(x)$ for all $g \in G$ and $x \in \mathcal{X}$.\footnote{
Note that $g$ acts on $\mathcal{X}$ and $\mathcal{Y}$ with different representations.
For instance, actions on $\mathcal{X}$ might be rotations, while actions on $\mathcal{Y}$ might be shifts.
}
In classification, the group acts trivially on labels, so equivariance reduces to $G$-invariance:
$f(g \cdot x) = f(x) \quad \forall\, g \in G,\; x \in \mathcal{X}$.
A $G$-invariant classifier assigns a single label to every orbit, which is precisely the robustness property we seek.
We assume $f$ to be any classifier that is not already $G$-invariant, i.e.\ whose predictions can change under transformations in $G$, and we aim to render it $G$-invariant at test time.
We do this by introducing a $G$-invariant function $h: \mathcal{X} \to \mathcal{X}$ such that $f(h(gx)) = f(h(x))$.
We gain $G$-invariance without restricting $f$.
This process is called \emph{canonicalization} \citep{Kaba2023}, where $h$ maps input data to a specific orbit member $x \in Gx$, called the \emph{canonical form} of $x$.

\begin{definition}[Orbit Canonicalization \cite{Dym2024}]
    A canonicalization is a function $h: \mathcal{X} \to \mathcal{X}$ fulfilling the following invariance property for all $g \in G$ and $x \in \mathcal{X}$:
    $ h(g \cdot x) = h(x) \quad \text{and} \quad h(x) \in Gx. $
    \label{defn:Canonicalization}
\end{definition}

As there is no trivial choice for $h(x)$, the most popular approaches \cite{Kaba2023,Mondal2023} use optimization to find the canonical form of $x$.
In this setting, each $g \cdot x \in Gx$ is assigned an energy and an optimizer seeks to find the point of least energy $g^*$, corresponding to the canonical form.
The choice of energy function $E: \mathcal{X} \to \mathbb{R}$ is essential, as its minima should correspond to the canonical form of $x$ and should be unique up to symmetries.
The resulting $(g^*)^{-1} \cdot x$ is ``undoing'' the transformation applied to $x$, mapping it to its canonical form.
Such that
\begin{equation}
\label{eq:optimization_kaba}
    h(x) = (g^*)^{-1} \cdot x
    \quad \text{with} \quad
    g^* = \arg\min_{g \in G} E(g^{-1} \cdot x).
\end{equation}
For small discrete groups, this minimization can be done exactly. 
Continuous groups require optimization methods that find approximate solutions.
In principle $E$ can be any function. 
What it does not guarantee is a unique minimizer per orbit: a generic OOD score may attain its minimum at several transformations that are not related by the stabilizer, in which case the canonical form is not well-defined and consistency across the orbit can fail~\citep{Kaba2023}. 
We nonetheless forgo this formal uniqueness guarantee, since suboptimal solutions without such guarantees often perform well in practice~\citep{ITS}, and we verify the resulting canonical forms empirically.

\section{OOD-based Canonicalization}
\label{sec:method}

\subsection{Searching for Canonical Forms}
An OOD score $S: \mathcal{X} \to \mathbb{R}$ measures alignment with the training distribution, with lower values indicating that $x$ is drawn from the distribution the classifier $f$ was trained on.
Our central insight is that, since canonical representatives reside near the data manifold~\cite{Singhal2025}, OOD scores are natural candidates for the energy $E$ in \Cref{eq:optimization_kaba}.
This is a necessary but not sufficient condition.
An OOD score rewards transformations that look typical, which may differ from the correct semantic pose for classes with inherent pose ambiguity. 
Whether it is sufficient in practice is the empirical question we study.
We set $E(x) = S(x)$ and aim to find the lowest OOD score over all target transformations $G$.
Throughout, we assume the transformation group $G$ is specified in advance and contains the transformations the input may have undergone. 
This means that $g^* \cdot x$ is the most ID point.
This connection opens the full landscape of OOD scores for canonical form estimation, whereas prior work was confined to very limited options \cite{Kaba2023,Mondal2023,ITS,Singhal2025}. 
Similarly, the choice of optimization strategy over $G$ has received little attention, with prior work largely restricted to gradient-based optimization \cite{Mondal2023} or tree-searches \cite{ITS} or Bayesian Optimization \cite{Singhal2025}.

In this work, we study various choices for algorithms to find the minimum in \Cref{eq:optimization_kaba}.
We focus on heuristics rather than optimal solvers as for continuous groups this would be intractable.
This leads to potentially suboptimal results (i.e., local minima $g$ with $S((g^*)^{-1} \cdot x) \leq S(g^{-1} \cdot x)$).
Even when $S$ is convex in $x$, it is generally non-convex as a function of $g$, necessitating multi-start optimization rather than a single descent run~\cite{Shumaylov2025}.
These local minima can yield predicted canonical forms that are themselves OOD~\citep{ITS,Singhal2025}.
This sub-optimality is particularly harmful when the found $g^{-1} \cdot x$ (local minima) is far from the true canonical form (global minima), degrading the classifier's accuracy on standard inputs \cite{ITS}.
We address this with a gating mechanism for canonicalization functions $h$, which we will introduce next.


\subsection{Gated Canonicalization}
A practical failure mode of continuous canonicalization is the degradation of accuracy on data that is already in-distribution (which we show in the experiments). 
Unnecessary transformations perturb features and lower downstream confidence.
To prevent this, we introduce \emph{Gated Canonicalization}, which comprises two gates: 
a selection gate $\alpha_\tau(x)$ and an acceptance gate $\alpha(x)$.
The former ensures that canonicalization is only applied if $x$ is OOD.
The latter ensures that the output of the canonicalization function reduces the OOD score, i.e., $S(\alpha_\tau(x)) \leq S(x)$.
This is equivalent to always including the identity in the search, though stronger acceptance criteria may break this equivalence (see \Cref{app:selectiongate}).
We define two canonicalization gates:
\begin{align} \label{eq:selection_acceptance}
    \text{selection gate} \quad
    \alpha_\tau(x) &= \begin{cases} (g^*)^{-1} \cdot x & \text{if } S(x) > \tau, \\ x & \text{otherwise,} \end{cases} \\
    \text{acceptance gate} \quad
    \alpha(x) &= \begin{cases} (g^*)^{-1} \cdot x & \text{if } S((g^*)^{-1} \cdot x) < S(x), \\ x & \text{otherwise,} \end{cases}
\end{align}
where $\tau \in \mathbb{R}$ is a user-defined OOD threshold. 
In practice, $\tau$ is selected on a held-out validation set augmented with the same transformations, requiring no labels for the transformed test data.
We study its effect in \Cref{sec:experiments} and report validation-selected operating points in \Cref{app:selectiongate}.
These mechanisms decouple the task of ID preservation from OOD robustness, treating the canonicalization process as a conditional corrective mapping.
A visual overview is illustrated in \Cref{fig:pitch}.

\section{OOD Scores and Search Methods}

\newcommand{\timebar}[2][cyan!40!white]{%
  \pgfmathsetmacro{\bh}{#2 / 100 * 1.8}%
  \begin{tikzpicture}[baseline=0pt, inner sep=0pt, outer sep=0pt]
    \path[use as bounding box] (0, 0) rectangle (0.25cm, -1.8cm);
    \fill[#1] (0, 0) rectangle (0.25cm, -\bh cm);
    \node[black, rotate=90, font=\tiny\bfseries, inner sep=0pt,
          anchor=north east]
      at (0.125cm, 0) {#2\%};
  \end{tikzpicture}%
}

\begin{table}[t]
\centering
\setlength{\tabcolsep}{1pt}
\caption{
Tested OOD scores, grouped by family (logit-based, activation-rectification, prototype-based, kNN-based; separated by rules). 
For each score we mark whether it requires stored training samples (TS), hyperparameter tuning (HP), or an auxiliary logit-based score (AS). 
The bottom row visualizes the relative inference-time increase (RT) of the full search relative to the Energy score, with bar height proportional to the increase.
Bars below the axis indicate scores faster than Energy. 
Values are averaged across MNIST, EMNIST, TU Berlin and ModelNet10.
Exact numbers are given in \Cref{tab:speed_comparison}.}
\label{tab:ood_overview}
\begin{tabular}{l *{20}{c}}
& \rotatebox{75}{Energy}
& \rotatebox{75}{Entropy}
& \rotatebox{75}{GEN}
& \rotatebox{75}{MSP}
& \rotatebox{75}{MaxLogit}
& \rotatebox{75}{OpenMax}
& \rotatebox{75}{VIM}
& \rotatebox{75}{Laplace}
& \rotatebox{75}{ReAct}
& \rotatebox{75}{ASH}
& \rotatebox{75}{DICE}
& \rotatebox{75}{MD}
& \rotatebox{75}{RMD}
& \rotatebox{75}{SHE}
& \rotatebox{75}{Proto}
& \rotatebox{75}{PC-Proto}
& \rotatebox{75}{kNN}
& \rotatebox{75}{Trust Score}
& \rotatebox{75}{kNN Mix}
& \rotatebox{75}{PC-kNN Mix} \\
\cmidrule(lr){2-6}\cmidrule(lr){7-9}\cmidrule(lr){10-12}
\cmidrule(lr){13-17}\cmidrule(lr){18-21}
TS
  & -- & -- & -- & -- & --          
  & \checkmark & \checkmark & \checkmark   
  & \checkmark & -- & \checkmark    
  & \checkmark & \checkmark & \checkmark & \checkmark & \checkmark  
  & \checkmark & \checkmark & \checkmark & \checkmark \\            
HP
  & -- & -- & \checkmark & -- & --
  & \checkmark & \checkmark & \checkmark
  & \checkmark & \checkmark & \checkmark
  & \checkmark & \checkmark & \checkmark & \checkmark & \checkmark
  & \checkmark & \checkmark & \checkmark & \checkmark \\
AS
  & -- & -- & -- & -- & -- & -- & --
  & \checkmark & \checkmark & \checkmark & \checkmark
  & -- & -- & -- & -- & -- & -- & -- & -- & -- \\
\midrule
RT
  & \timebar{0.0}    
  & \timebar{-3.0}   
  & \timebar{-1.5}   
  & \timebar{-3.5}   
  & \timebar{-1.7}   
  & \timebar{32.5}   
  & \timebar{1.8}    
  & \timebar{53.9}   
  & \timebar{-1.0}   
  & \timebar{2.2}    
  & \timebar{6.0}    
  & \timebar{13.3}   
  & \timebar{7.8}    
  & \timebar{2.2}    
  & \timebar{3.8}    
  & \timebar{9.3}    
  & \timebar{14.3}   
  & \timebar{52.8}   
  & \timebar{37.5}   
  & \timebar{43.5} \\
\vspace{-1.5cm} 
\end{tabular}
\end{table}

\paragraph{OOD Scoring Functions.}
We limit our focus to training-free OOD scores that require no architectural changes. Our central premise is that canonical representatives lie near the data manifold, so any reliable OOD score is a natural energy function for canonicalization. We evaluate four families of detectors.
\emph{Logit-based methods} operate on classifier outputs and require no stored features. We include Maximum Softmax Probability (MSP) \citep{HendrycksOOD2022}, MaxLogit \citep{Hendrycks2022}, Energy \citep{Liu2020}, Entropy, Generalized Entropy (GEN) \citep{Liu2023}, OpenMax \citep{Bendale2016}, Virtual-logit Matching (VIM) \citep{wangViMOutOfDistributionVirtuallogit2022}, and the Laplace Approximation \citep{Daxberger2021} with energy, entropy, and mutual information (MI) scores \citep{TorchUncertainty}.
\emph{Activation-rectification methods} modify intermediate features before computing a logit-based score, amplifying distributional differences. We include Rectified Activations (ReAct) \citep{ReAct2021}, Activation Shaping (ASH) \citep{Djurisic2023}, and Directed Sparsification (DICE) \citep{DICE}.
\emph{Prototype-based methods} measure distance to class representatives in feature space. We include Mahalanobis Distance (MD) \citep{Lee2018}, Relative Mahalanobis Distance (RMD) \citep{Ren2021}, Simplified Hopfield Energy (SHE) \citep{Zhang2023}, and prototype distances using Euclidean, cosine, or a convex combination of both, applied either globally or per predicted class (PC-Proto).
\emph{kNN-based methods} score inputs by their average distance to the $k$ nearest in-distribution embeddings \citep{Sun2022}. We include variants using Euclidean and cosine distance, a convex mixture (kNN Mix), per-class versions (PC-kNN, PC-kNN Mix), and Trust Score \citep{Jiang2018}. These methods require storing training embeddings but no additional training.
Table~\ref{tab:ood_overview} summarizes the requirements of each detector. Further implementation details are in Appendix~\ref{app:ood}.

\begin{table}[t]
\caption{
Tested search algorithms. 
\textit{(left)} For each method we mark whether it is a global search (GS), gradient-based (GB), and parallelizable (PL), and visualize the relative inference-time increase (RT) over Random Search as a bar (height proportional to the increase, averaged across MNIST, EMNIST, TU Berlin and ModelNet10). 
\textit{(right)} Worst-case time complexity in Big-O.}
\label{tab:optimizer_overview}
\vspace{2pt} 
\begin{minipage}[t]{0.4\linewidth}
  \centering
  \setlength{\tabcolsep}{1.pt}
  \begin{tabular}{l *{9}{c}}
  & \rotatebox{75}{RS}
  & \rotatebox{75}{RS-LR}
  & \rotatebox{75}{SA}
  & \rotatebox{75}{GD}
  & \rotatebox{75}{CD-S}
  & \rotatebox{75}{CD}
  & \rotatebox{75}{WCD}
  & \rotatebox{75}{WCD-L}
  & \rotatebox{75}{ITS} \\
  \cmidrule(lr){2-5}\cmidrule(lr){6-10}
  GS
    & \checkmark & \checkmark & \checkmark & \checkmark
    & -- & -- & -- & -- & -- \\
  GB
    & -- & \checkmark & -- & \checkmark
    & -- & -- & --  & -- & -- \\
  PL
    & \checkmark & \checkmark & \checkmark & \checkmark
    & -- & -- & -- & --  & -- \\
  \midrule
  RT
    & \timebar{0.0}
    & \timebar{5.7}
    & \timebar{34.8}
    & \timebar{6.7}
    & \timebar{11.2}
    & \timebar{13.1}
    & \timebar{19.3}
    & \timebar{27.3}
    & \timebar{24.76} \\
  \end{tabular}
\end{minipage}
\hfill
\begin{minipage}[t]{0.55\linewidth}
  \centering
  \setlength{\tabcolsep}{2.5pt}
  \begin{minipage}[t]{0.45\linewidth}
  \vspace{-1.6cm}
    \begin{tabular}{l l}
    Method & Complexity \\
    \midrule
    RS       & $\mathcal{O}(N)$         \\
    RS-LR    & $\mathcal{O}(N + Lk)$    \\
    SA       & $\mathcal{O}(Rk)$        \\
    GD       & $\mathcal{O}(Rk)$        \\
    WCD      & $\mathcal{O}(M(d{+}W)c)$ \\
    ITS      & $\mathcal{O}(MHd)$       \\
    \end{tabular}
  \end{minipage}%
  \hfill
  \begin{minipage}[t]{0.44\linewidth}
    \vspace{-2.1cm}
    \raggedright\small
    $N$: samples\\
    $L$: local pts\\
    $k$: iterations\\
    $R$: runs/restarts\\
    $M$: samples/dim\\
    $d$: dims\\
    $W$: first-dim weight\\
    $c$: cycles\\
    $H$: hypotheses
  \end{minipage}
\end{minipage}
\vspace{-1.2cm} 
\end{table}

\paragraph{Search Algorithms.}
We evaluate algorithms covering a spectrum from global exploration to local exploitation.
\emph{Random Search (RS)} \citep{Sobol1967} samples the parameter space via Sobol sequences for uniform coverage. \emph{RS with Local Refinement (RS-LR)} applies Adam-based gradient descent to the best candidate from global sampling. \emph{Simulated Annealing (SA)} \citep{Kirkpatrick1983} uses neighborhood sampling \citep{Spall2003} with multiple parallel runs to reduce latency. \emph{Multi-start Gradient Descent (GD)}, as used in \citep{Shumaylov2025}, runs Adam from several random initializations.
As a non-global baseline inspired by ITS \citep{ITS}, we study \emph{Coordinate Descent (CD)}, which sequentially optimizes each transformation parameter via uniform random sampling, iterating over multiple cycles. Variants include \emph{CD-S} (one cycle), \emph{Weighted CD (WCD)} (increased sampling density in the first dimension), and \emph{WCD-L} (equally spaced rather than random samples). Table~\ref{tab:optimizer_overview} summarizes properties and complexities. Full details are in Appendix~\ref{app:search}.

\section{Experiments}
\label{sec:experiments}

We benchmark our canonicalization framework against competitive baselines in \Cref{sec:exp:benchmark}, ablate the choice of OOD score in \Cref{sec:exp:ood} and search algorithm in \Cref{sec:exp:search}.
In \Cref{sec:exp:gating}, we add our gating mechanism and study its impact.
All our experiments are conducted with lightweight compute hardware (NVIDIA RTX 5080, Intel 13600k). 
We make our source code publicly available for the reproducibility of our experiments. 
We refer the reader to \Cref{app:implementation} or our source code for more details on the used hyperparameters.

\begin{tcolorbox}[colback=white, colframe=black, boxrule=0.5pt, 
                  left=4pt, right=4pt, top=3pt, bottom=3pt]
\small\textbf{Finding.}
Across all benchmarks, random search with local refinement (\textbf{RS-LR}) combined with a distance-based OOD score performs best (\Cref{sec:exp:search,sec:exp:ood}). The single best score is dataset dependent but always distance-based: kNN for MNIST, PC-kNN for EMNIST, kNN Mix for ModelNet10, and PC-Proto for TU Berlin.
On the validation set, the exact same detectors achieve the top scores across all datasets except ModelNet10, where \textbf{PC-kNN Mix} performs best. We refer to using the best configuration(RS-LR plus score) on the validation set as \textbf{OODC} (OOD-based Canonicalization) and use it throughout unless stated otherwise.
\end{tcolorbox}

\paragraph{Benchmarks and Preprocessing}
We conducted experiments on MNIST, EMNIST Balanced~\citep{EMNIST} and SI-Score (rotation) ~\citep{SIScore} as 2D image classification datasets.
SI-Score~\citep{SIScore} is a robustness benchmark, featuring ImageNet \cite{ImageNet} foreground objects randomly transformed onto different backgrounds.
We further included TU Berlin~\citep{Eitz2012} as a 2D sketch dataset drawn with one line and ModelNet10~\citep{Wu2015} as a 3D point cloud dataset.
We created additional test sets by augmenting the existing test sets of MNIST, EMNIST and TU Berlin by rotation (angles uniformly from $[0,2\pi)$), shear (with factors from $[-0.5,0.5]$), and scaling (from $[1/1.8,1.8]$) with bicubic interpolation during resampling.\footnote{TU Berlin is converted to sequences $(x, y, \text{pen state})$ by keeping edge points, as this data can be transformed without any interpolation artifacts.}
For ModelNet10 only rotation is applied as scale and translation are normalized a priori.
In \Cref{app:implementation} we provide further implementation details. The search uses the same transformations that are used for augmentation of the dataset.

\paragraph{Downstream Models and Metrics}
For MNIST and EMNIST we use a small residual network, for TU Berlin a BiLSTM~\citep{LSTM}, for ModelNet10 a PointNet++~\citep{PointNetPP}, and for SI-Score a ViT-B16 \cite{ViT} and a ResNet50 \cite{ResNets}. 
We provide further details on training and architectures in \Cref{app:architecture_training}.
We report the average top-1 test accuracies (mean over multiple runs with 1.96 × standard error to indicate statistical significance).
We also want to report how close we get to the canonical form (i.e., the energetic minimum).
As the distance between the predicted and the ground truth affine transformation matrix is degenerate due to its non-trivial equivalence classes, we use a score-based distance.
More specifically, we measure the relative distance between OOD scores using 
\begin{equation}
\label{eq:relative_distance}
\text{Scoring Distance (Err.):} \quad
\frac{
S((g^*)^{-1}gx) - S((g^*_{\min})^{-1}gx)
}{
S((g^*_{\max})^{-1}gx) - S((g^*_{\min})^{-1}gx)
}
\in [0,1].
\end{equation}
where $g^*$ is the found solution, $gx$ the transformed input, $g^*_{\min}$ and $g^*_{\max}$ the transformations with minimal and maximal score across all runs and methods.

\subsection{How competitive is our OOD-based canonicalization?}
\label{sec:exp:benchmark}

\begin{table}[t]
\centering
\caption{Accuracy (\%) on the original (top row per dataset) and transformed (bottom row per dataset) test sets. 
Methods are grouped into those that require retraining and those that do not. 
Retraining baselines: augmented training (Augm.), an equivariant SteerableCNN~\cite{SteerableCNNs} or Vector Neuron model~\cite{Deng_2021_ICCV} (Equiv.), a spatial transformer trained on augmented data (STN), and canonicalization with a learned energy model (Learned, \Cref{app:supervised}). 
Training-free methods: the vanilla classifier, Test-time Augmentation (TTA)~\cite{TIPooling}, vanilla ITS~\cite{ITS}, ITS with its logit energy (ITS\textsuperscript{a}), ITS with our best OOD score (ITS\textsuperscript{b}), and OODC. 
The best result within each group is shown in bold per row.
}
\label{tab:combined}
\begin{tabular}{l|cccc|ccccccc}
\toprule
& \multicolumn{4}{c|}{Requires Retraining} & \multicolumn{6}{c}{No Retraining} \\
Dataset & Augm & Equiv. & STN & Learned & Vanilla & TTA & ITS & ITS\textsuperscript{a} & \textbf{ITS\textsuperscript{b}} & \textbf{OODC} \\
\midrule
\multirow{2}{*}{MNIST}
  & 99.2 & \textbf{99.3} & \textbf{99.3} & 98.6 $\pm$ 0.1 & \textbf{99.6} & 83.0 & 76.3 & 93.8 & 98.1 & 95.1 $\pm$ 0.1 \\
  & 98.7 & \textbf{99.2} & \textbf{99.2} & 98.1 $\pm$ 0.1 & 37.7 & 49.8 & 60.4 & 75.5 & 85.0 & \textbf{89.8 $\pm$ 0.3} \\
\midrule
\multirow{2}{*}{EMNIST}
  & 83.2 & \textbf{84.1} & 83.6 & 76.7 $\pm$ 0.2 & \textbf{88.4} & 44.2 & 54.1 & 71.2 & 78.0 & 64.4 $\pm$ 0.2 \\
  & 82.1 & \textbf{83.9} & 82.8 & 75.6 $\pm$ 0.1 & 17.1 & 17.2 & 33.4 & 46.6 & 51.4 & \textbf{56.1 $\pm$ 0.4} \\
\midrule
\multirow{2}{*}{TU Berlin}
  & \textbf{42.4} & --- & 30.5 & 35.0 $\pm$ 0.8 & \textbf{51.6} & 18.9 & 36.6 & 39.7 & 40.5 & 34.0 $\pm$ 0.4 \\
  & \textbf{40.7} & --- & 28.5 & 31.2 $\pm$ 1.4 &  8.2 &  8.2 & 19.7 & 22.0 & 25.7 & \textbf{28.1 $\pm$ 0.3} \\
\midrule
\multirow{2}{*}{ModelNet10}
  & 80.1 & \textbf{87.8}  & 82.5 & 70.1 $\pm$ 0.5 & \textbf{91.2} & 47.7 & 33.6 & 32.4 & 86.7 & 60.6 $\pm$ 0.7 \\
  & 83.9 & \textbf{87.8} & 82.9 & 68.2 $\pm$ 1.6 & 20.5 & 28.3 & 27.1 & 27.6 & 54.0 & \textbf{59.6 $\pm$ 0.9} \\
\bottomrule
\end{tabular}
\end{table}

\begin{figure}[t]
    \centering
    \begin{minipage}[c]{0.37\textwidth}
        \centering
        \includegraphics[width=\linewidth]{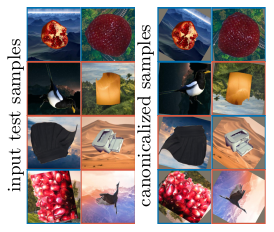}
    \end{minipage}
    \hfill
    \begin{minipage}[c]{0.62\textwidth}
        \centering
        \begin{tabular}{c p{0.4\linewidth} cc}
            \toprule
            & \textbf{Method} & \textbf{ViT-B16} & \textbf{ResNet50} \\
            \midrule
            \multirow{6}{*}{\rotatebox[origin=c]{90}{\tiny Baselines}} 
            & Vanilla Classifier & 38.8 & 48.0 \\
            & RS Augm.~\citep{ITS} & 41.0 & -- \\
            & Chefer et al.~\citep{Chefer2022} & 44.8 & -- \\
            & Chefer et al. + AS~\citep{Chefer2022} & 46.2 & -- \\
            & ITS~\citep{ITS} & 49.0 & -- \\
            & TTA \cite{TIPooling} & 49.5 & 54.9 \\
            \midrule
            \multirow{2}{*}{\rotatebox[origin=c]{90}{\tiny Score}} 
            & Energy of Logits & 51.4 $\pm$ 0.2 & 48.3 $\pm$ 0.2 \\
            & \textbf{PC-kNN} & \textbf{57.9 $\pm$ 0.1} & \textbf{60.7 $\pm$ 0.1} \\
            \bottomrule
        \end{tabular}
    \end{minipage}
    \caption{
Performance on the SI-Score (rotation) with canonicalized images on the left (blue correct, red incorrect classification), and test accuracies on the right.
    }
    \label{fig:sicore}
\end{figure}

\begin{table}[t]
    \centering
    \caption{Comparison of search strategies on different datasets with 60 evaluations, showing random search with local refinement performing best overall.}
    \label{tab:search_strat}
    \begin{tabular}{lrrrrrrrr}
    \toprule
     Method & \multicolumn{2}{r}{MNIST} & \multicolumn{2}{r}{EMNIST} & \multicolumn{2}{r}{ModelNet10} & \multicolumn{2}{r}{TU Berlin} \\
     & Acc $\uparrow$ & Err $\downarrow$ & Acc $\uparrow$ & Err $\downarrow$ & Acc $\uparrow$ & Err $\downarrow$ & Acc $\uparrow$ & Err $\downarrow$ \\
     &  &  &  &  &  &  &  &  \\
    \midrule
    CD & $52.2 \pm .2$ & $0.42$ & $80.2 \pm .1$ & $0.33$ & $37.5 \pm .4$ & $0.52$ & $24.4 \pm .3$ & $0.37$ \\
    CD-S & $52.2 \pm .1$ & $0.42$ & $80.1 \pm .2$ & $0.33$ & $36.0 \pm .3$ & $0.59$ & $24.5 \pm .3$ & $0.37$ \\
    GD & $54.0 \pm .2$ & $0.57$ & $85.6 \pm .2$ & $0.42$ & $39.6 \pm .5$ & $0.55$ & --- &  \\
    ITS & $55.7 \pm .1$ & $0.44$ & $83.6 \pm .1$ & $0.34$ & $39.2 \pm .1$ & $0.59$ & $25.3 \pm .1$ & $\mathbf{0.34}$ \\
    SA & $56.3 \pm .2$ & $0.56$ & $85.0 \pm .3$ & $0.47$ & $40.2 \pm .4$ & $0.54$ & $21.3 \pm .3$ & $0.47$ \\
    RS & $57.4 \pm .2$ & $0.56$ & $87.0 \pm .2$ & $0.46$ & $40.3 \pm .4$ & $0.53$ & $21.1 \pm .3$ & $0.47$ \\
    RS-LR & $\mathbf{59.9 \pm .2}$ & $0.36$ & $\mathbf{88.0 \pm .2}$ & $0.27$ & $\mathbf{41.9 \pm .4}$ & $\mathbf{0.44}$ & --- &  \\
    WCD & $56.0 \pm .2$ & $0.32$ & $82.6 \pm .2$ & $0.24$ & $38.4 \pm .5$ & $0.52$ & $26.1 \pm .2$ & $0.36$ \\
    WCD-L & $57.0 \pm .1$ & $\mathbf{0.30}$ & $83.0 \pm .1$ & $\mathbf{0.22}$ & $40.2 \pm .4$ & $0.48$ & $\mathbf{28.7 \pm .5}$ & $0.69$ \\
    \bottomrule
    \end{tabular}
\end{table}

We compared our OOD-based canonicalization framework on various benchmarks (including ID and OOD test sets) against multiple baselines.
Results for MNIST, EMNIST, TU Berlin, and ModelNet10 are shown in \Cref{tab:combined}. All retraining-based models were trained on augmented data. 
The STN \cite{STN} used a smaller version of the main model, with one-quarter of the channels. 
Test-time Augmentation averaged predictions over eight transformation samples, always including the identity.
Equivariant architectures perform best among approaches requiring retraining, while the vanilla model leads among training-free variants but lacks robustness.
OODC achieves the highest accuracy on transformed test data among all training-free methods.
On untransformed data, however, its accuracy falls well below the vanilla model.
This is inherent to always-on canonicalization: every input is transformed, so already-aligned inputs are needlessly perturbed and their features distorted.
The same effect appears for the STN, which also transforms unconditionally, indicating that the drop is the cost of unconditional robustness rather than a weakness of the OOD score.
In \Cref{sec:exp:gating} we show that gating restores most of the in-distribution accuracy while retaining the gains on transformed inputs, converting this apparent trade-off into a controllable one.

Next, we evaluate our canonicalization framework on SI-Score (Rotation) \cite{SIScore} as a large-scale robustness benchmark using a ResNet-50~\citep{ResNets} and a ViT-B16 \cite{ViT} backbone.
For optimization, we use 17 equally spaced samples as we only consider rotation.
Due to computational cost, we restricted OODC to PC-kNN here and additionally report the energy of the logits, without exploring other scores.
Hyperparameters were optimized on rotated ImageNet\citep{ImageNet} images (distinct from data used for embeddings).
For comparison, we include results from three baselines: ITS \citep{ITS} (using their logit-based energy detector and iterative search), foreground fine-tuning for ViT \cite{Chefer2022}, and Test Time Augmentation (TTA) over 16 equally spaced rotations.
The results in \Cref{fig:sicore} demonstrate that feature-based OOD scores significantly outperform previous approaches. 
On ResNet-50, OODC improves accuracy by over 12 points over the vanilla model, while logit energy yields a negligible change.

\begin{figure}[t]
    \centering
    \includegraphics[width=0.8\textwidth]{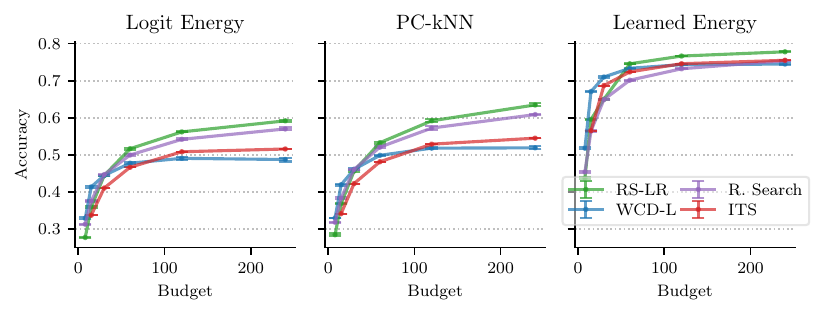}%
    \caption{Accuracy on EMNIST over different budget constraints.}%
    \label{fig:search_scaling}
\end{figure}

\begin{figure}[t]
    \centering
    \includegraphics[width=\textwidth]{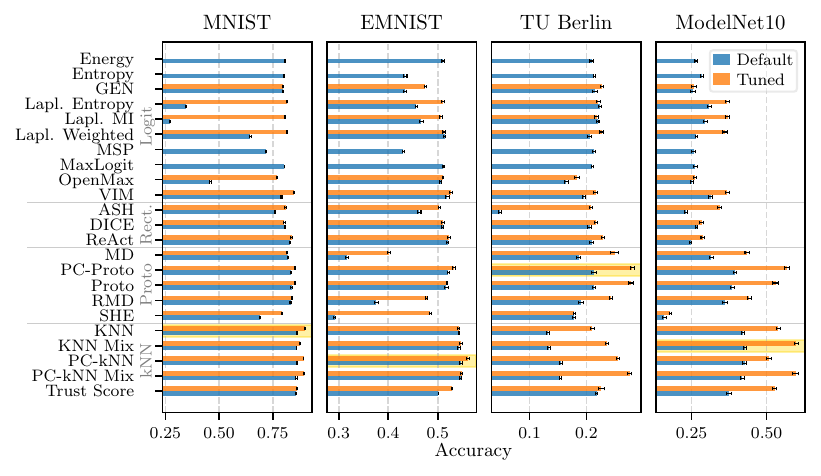}
    \caption{Comparison of unsupervised OOD scores across datasets with the best test score highlighted. Distance-based OOD scores perform best overall.}%
    \label{fig:comparision_detectors_1}
\end{figure}

\subsection{What Search Algorithm to choose for Canonicalization?}
\label{sec:exp:search}

As testing all classifiers and combinations is computationally not feasible, we opted to evaluate the search on several selected score functions first.
As scores for this experiment, we use the negative energy of the logits~\citep{Liu2020}, distance to the nearest embedding using cosine distance at the last layer (PC-kNN). 
In addition, we train a classifier to distinguish between data points that the downstream model predicts correctly and those it predicts incorrectly. 
We refer to this approach as a trained energy model.
The goal is to have a search algorithm that works with a wide range of possible scores, so we use the average accuracy across these scores as the metric.
A main limiting factor of the searches is the computation time which is mainly decided by the number of model calls the methods make.
For a fair comparison, the first experiment uses a fixed budget of 60 sample evaluations, chosen due to its divisibility.
A forward pass counts as one evaluation, while a forward and backward pass counts as two to reflect the higher cost.
The BiLSTM model requires unrolling, which is more expensive, so we do not apply gradient-based methods on the TU Berlin dataset.
The hyperparameters for each method and dataset are optimized for 30 iterations, using the average accuracy across all scores, while ensuring that the computational budget is not exceeded.
The results are shown in \Cref{tab:search_strat}.  The relative distance varied by at most 0.003 (1.96 × standard error), so it is omitted for readability.

We found random search outperforms most other search algorithms.
Coordinate descent-based methods find solutions closer to the minimum than random search, though often with lower accuracy, except on TU Berlin.
Using local refinement can boost the accuracy slightly and find results closer to the minima, but typically not as close as WCD-L. 
As accuracy is typically prioritized, this makes it the preferable method within this budget.
Next, we investigate if this holds for different budgets, re-optimizing hyperparameters for each.
The results for previously well-performing methods on EMNIST are shown in \Cref{fig:search_scaling} (see \Cref{app:budget} for the other benchmarks).
On most datasets, RS-LR performs well with budgets above 30, though coordinate descent-based methods may outperform it at lower budgets. We found TU Berlin to be an exception where ITS works better.
Overall, we recommend RS-LR for its simplicity, dimension-independent scaling, and strong performance at budgets where canonicalization is reliable.

\subsection{What OOD Scoring Function to choose for Canonicalization?}
\label{sec:exp:ood}

\begin{figure}[t]
\centering

\begin{subfigure}[t]{0.16\textwidth}
    \centering
    \includegraphics[width=\textwidth]{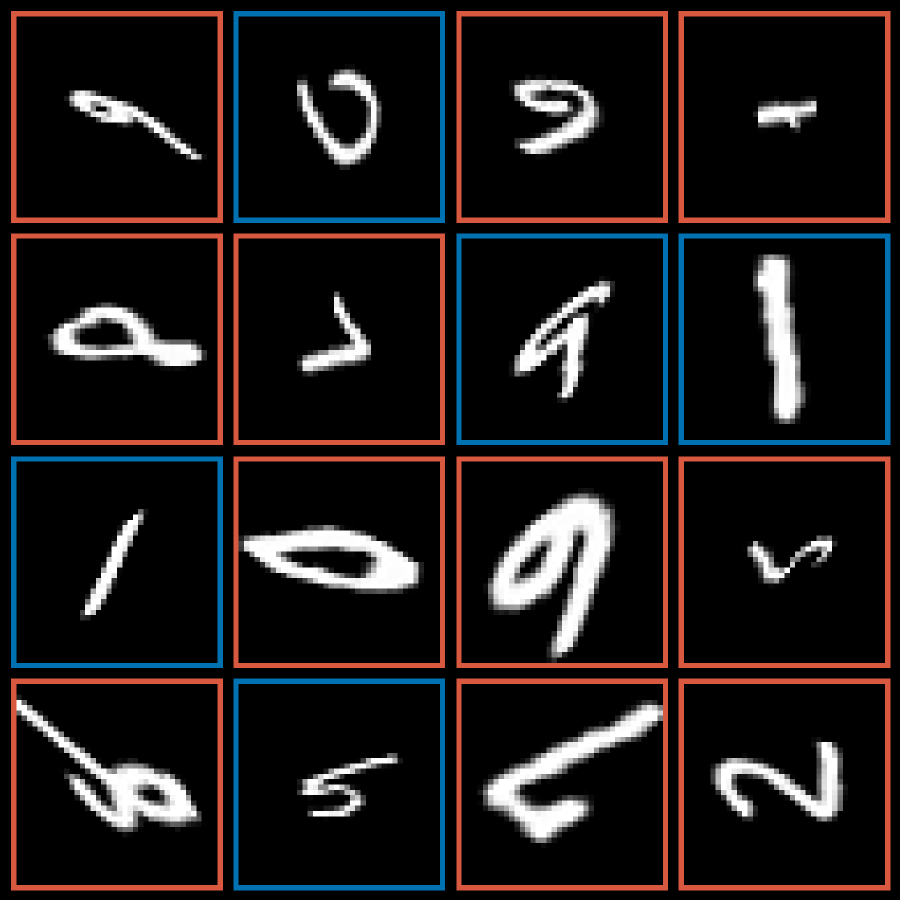}
    \caption{\centering MNIST Transformed}
\end{subfigure}
\begin{subfigure}[t]{0.16\textwidth}
    \centering
    \includegraphics[width=\textwidth]{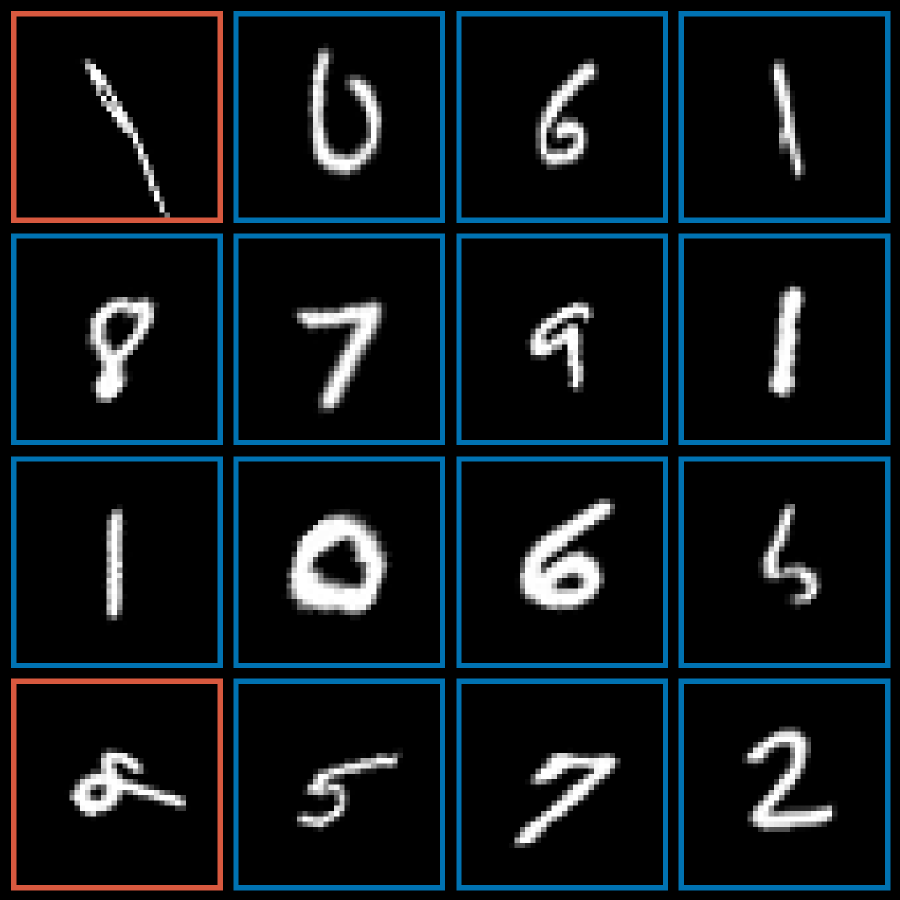}
    \caption{\centering MNIST Canon}
\end{subfigure}
\begin{subfigure}[t]{0.16\textwidth}
    \centering
    \includegraphics[width=\textwidth]{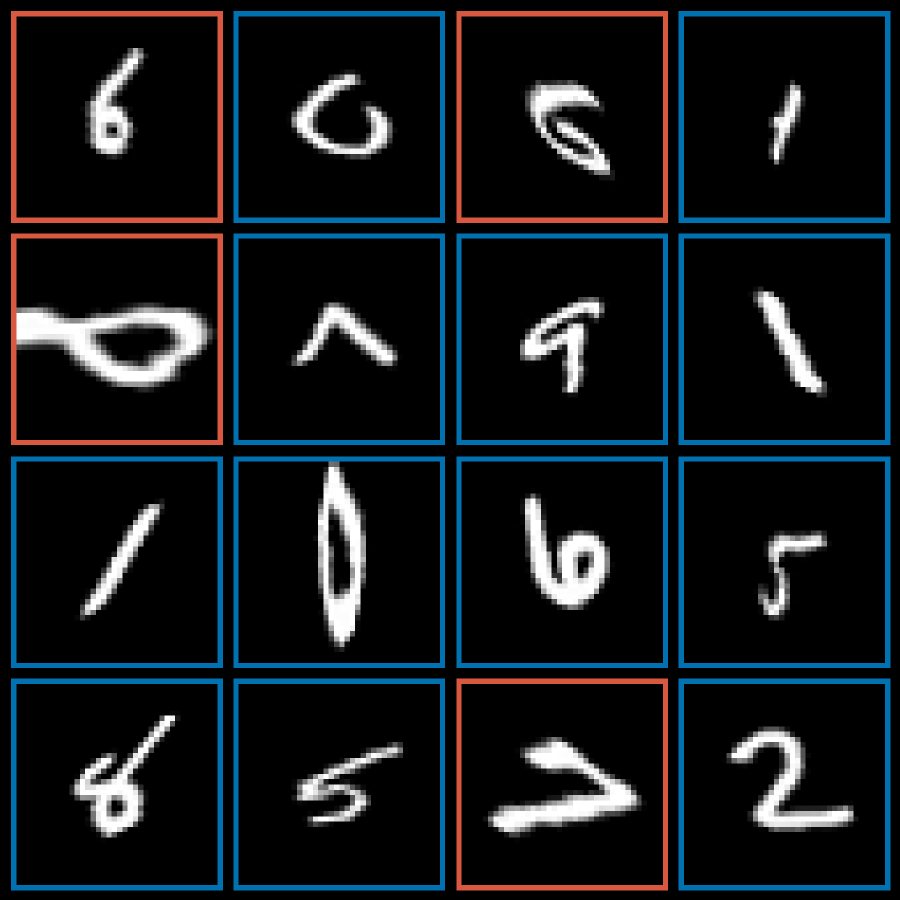}
    \caption{\centering MNIST ITS}
\end{subfigure}
\begin{subfigure}[t]{0.16\textwidth}
    \centering
    \includegraphics[width=\textwidth]{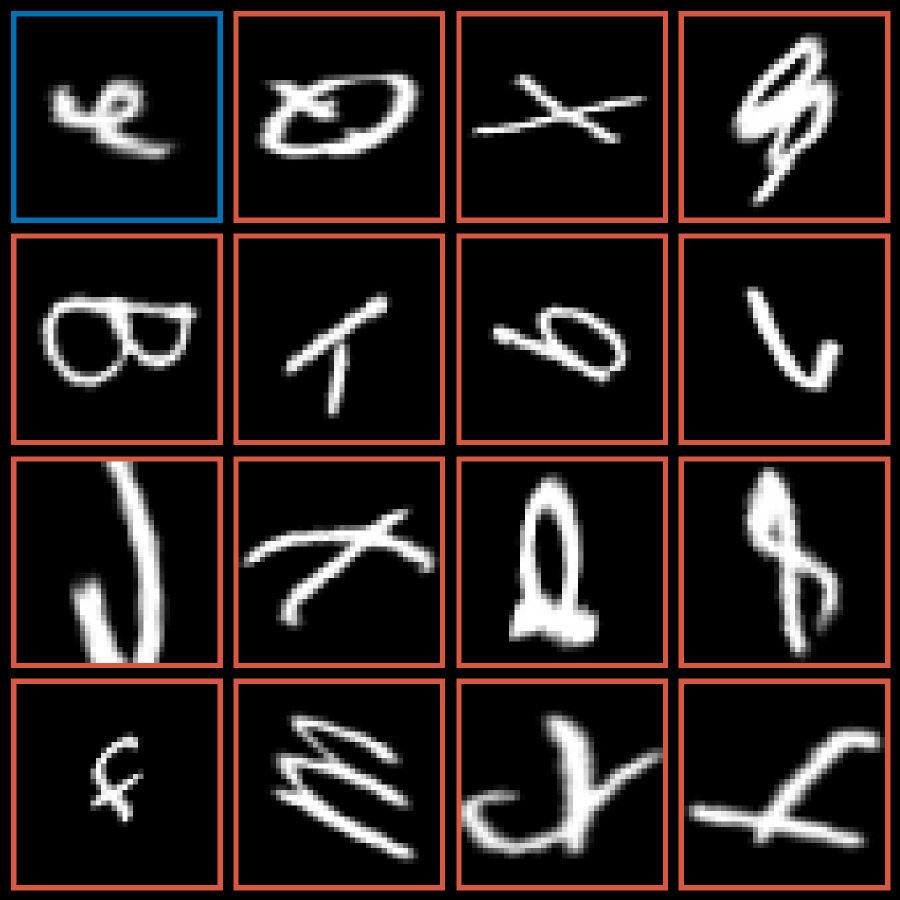}
    \caption{\centering EMNIST Transformed}
\end{subfigure}
\begin{subfigure}[t]{0.16\textwidth}
    \centering
    \includegraphics[width=\textwidth]{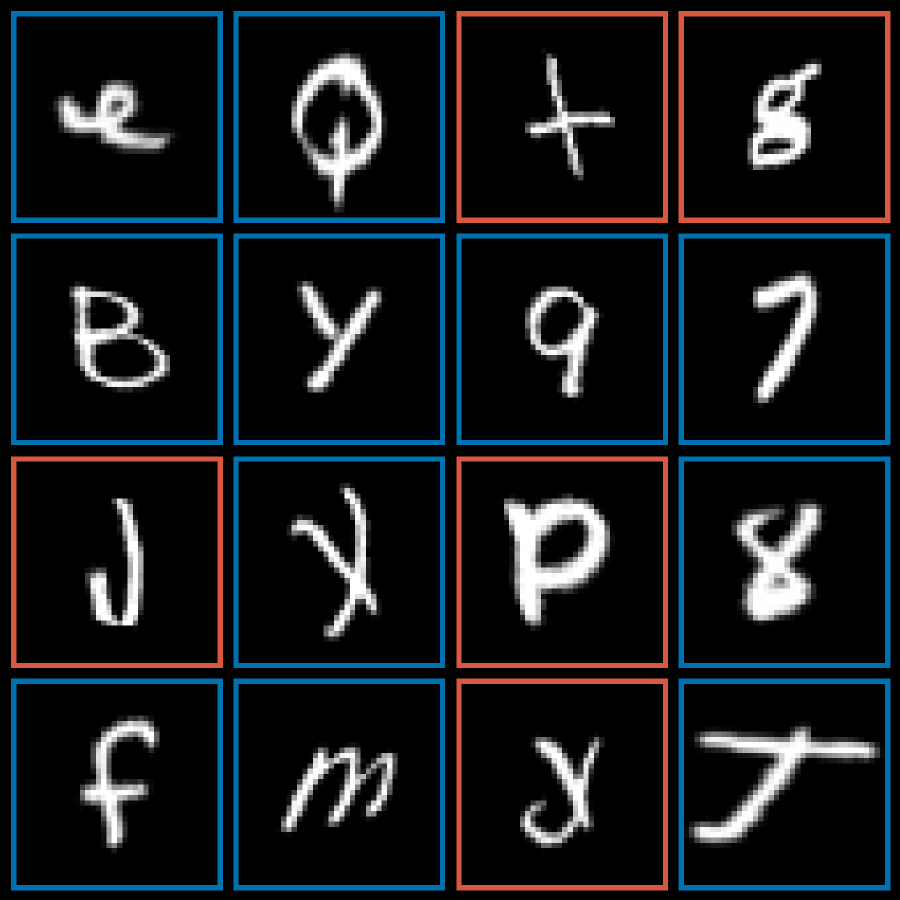}
    \caption{\centering EMNIST Canon}
\end{subfigure}
\begin{subfigure}[t]{0.16\textwidth}
    \centering
    \includegraphics[width=\textwidth]{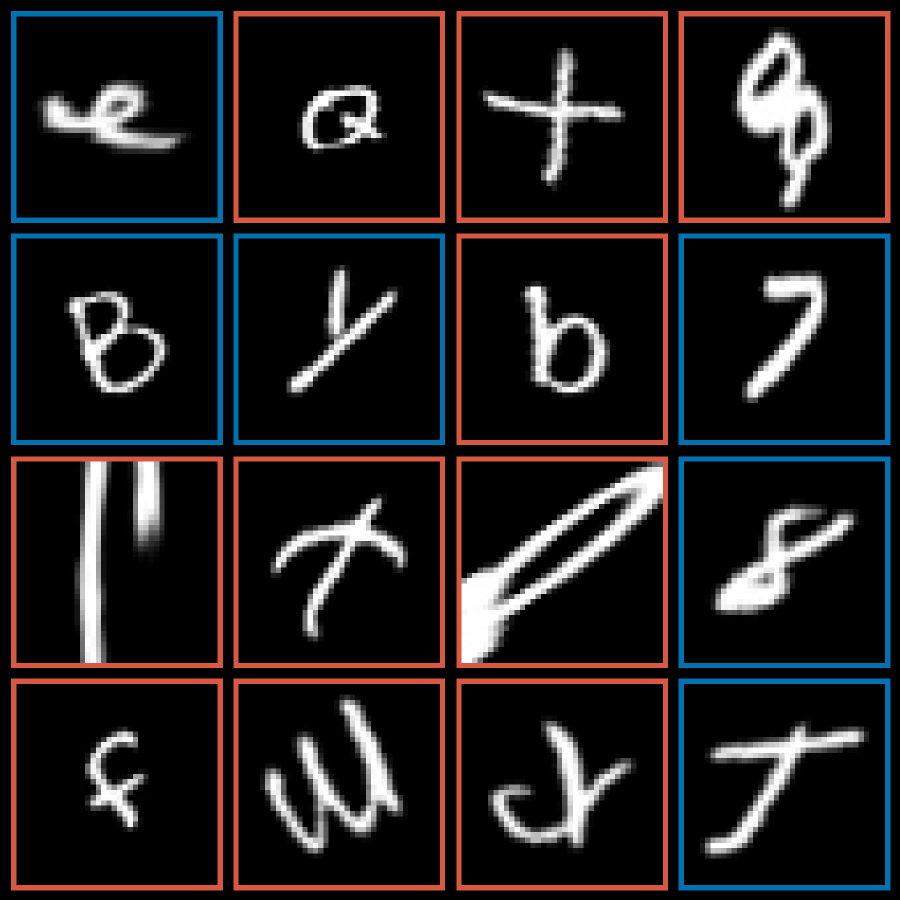}
    \caption{\centering EMNIST ITS}
\end{subfigure}

\begin{subfigure}[t]{0.16\textwidth}
    \centering
    \includegraphics[width=\textwidth]{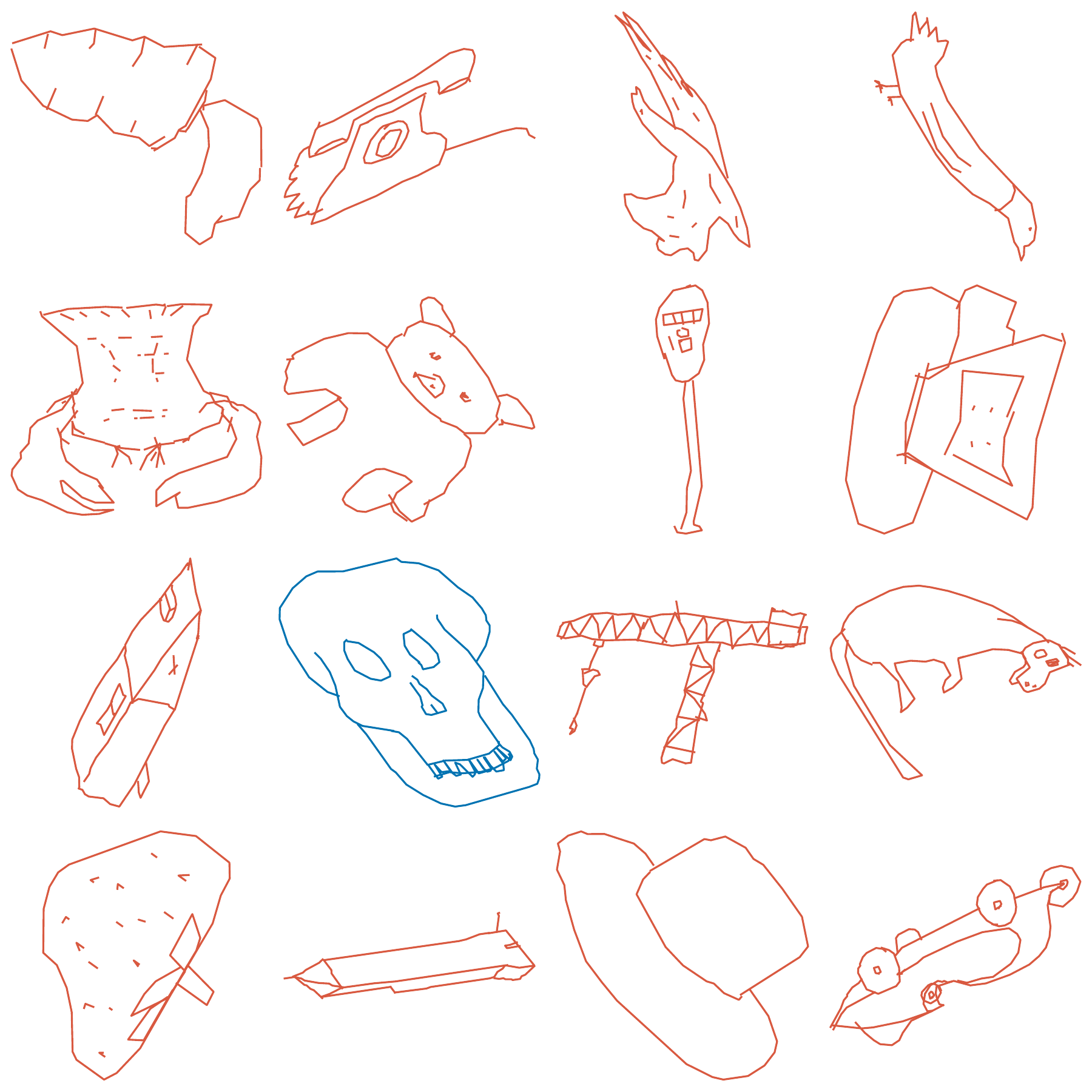}
    \caption{\centering TU Berlin Transformed}
\end{subfigure}
\begin{subfigure}[t]{0.16\textwidth}
    \centering
    \includegraphics[width=\textwidth]{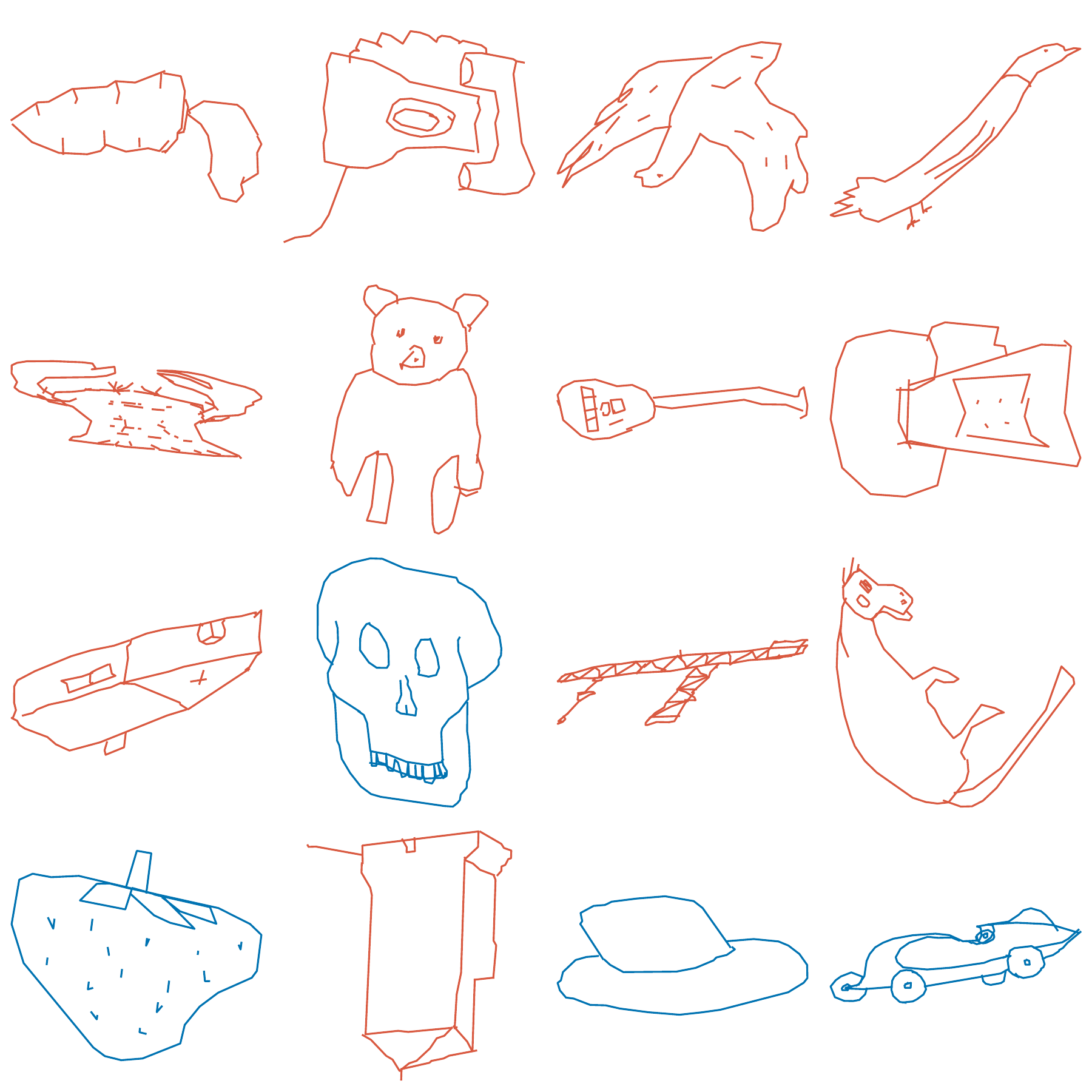}
    \caption{\centering TU Berlin Canon}
\end{subfigure}
\begin{subfigure}[t]{0.16\textwidth}
    \centering
    \includegraphics[width=\textwidth]{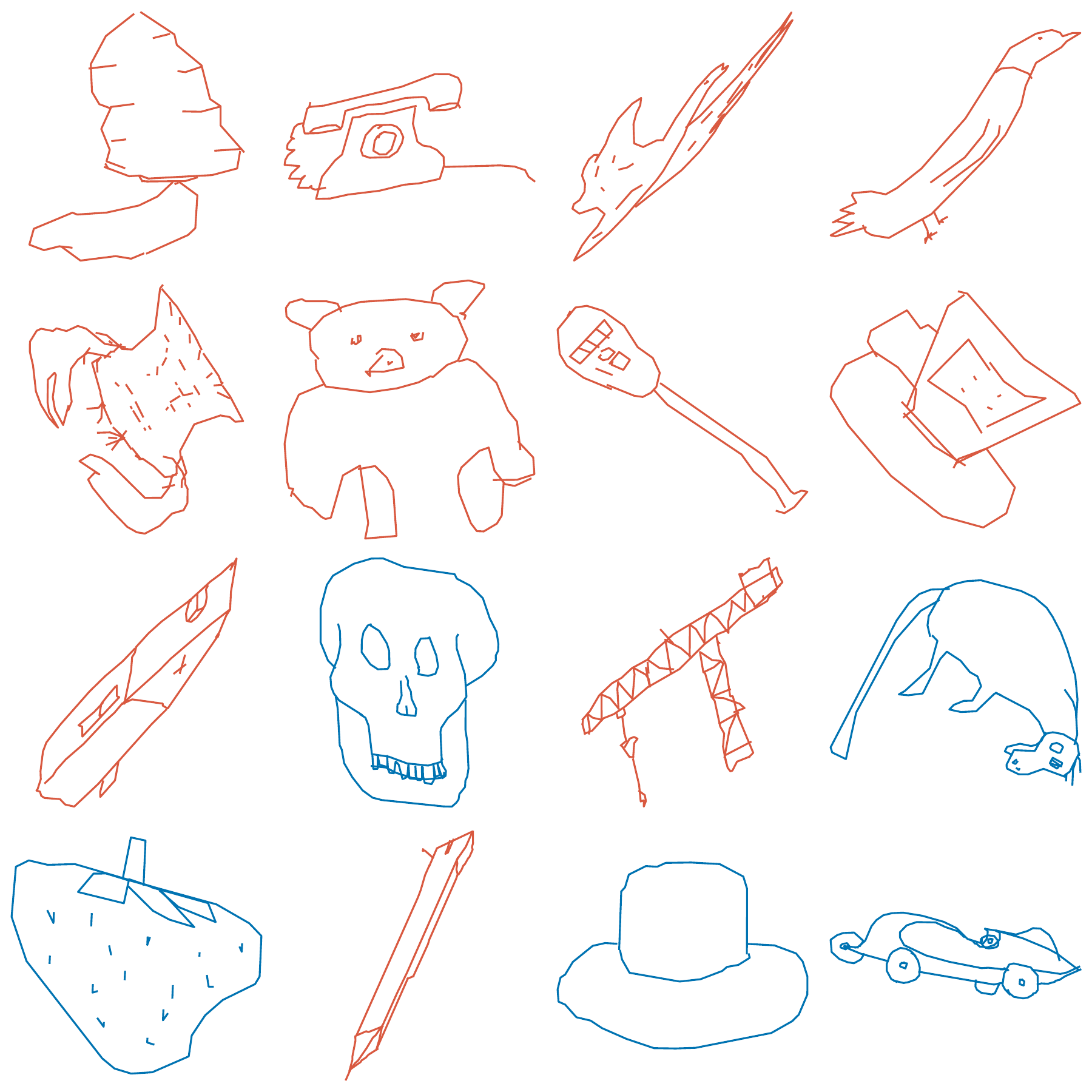}
    \caption{\centering TU Berlin ITS}
\end{subfigure}
\begin{subfigure}[t]{0.16\textwidth}
    \centering
    \includegraphics[width=\textwidth]{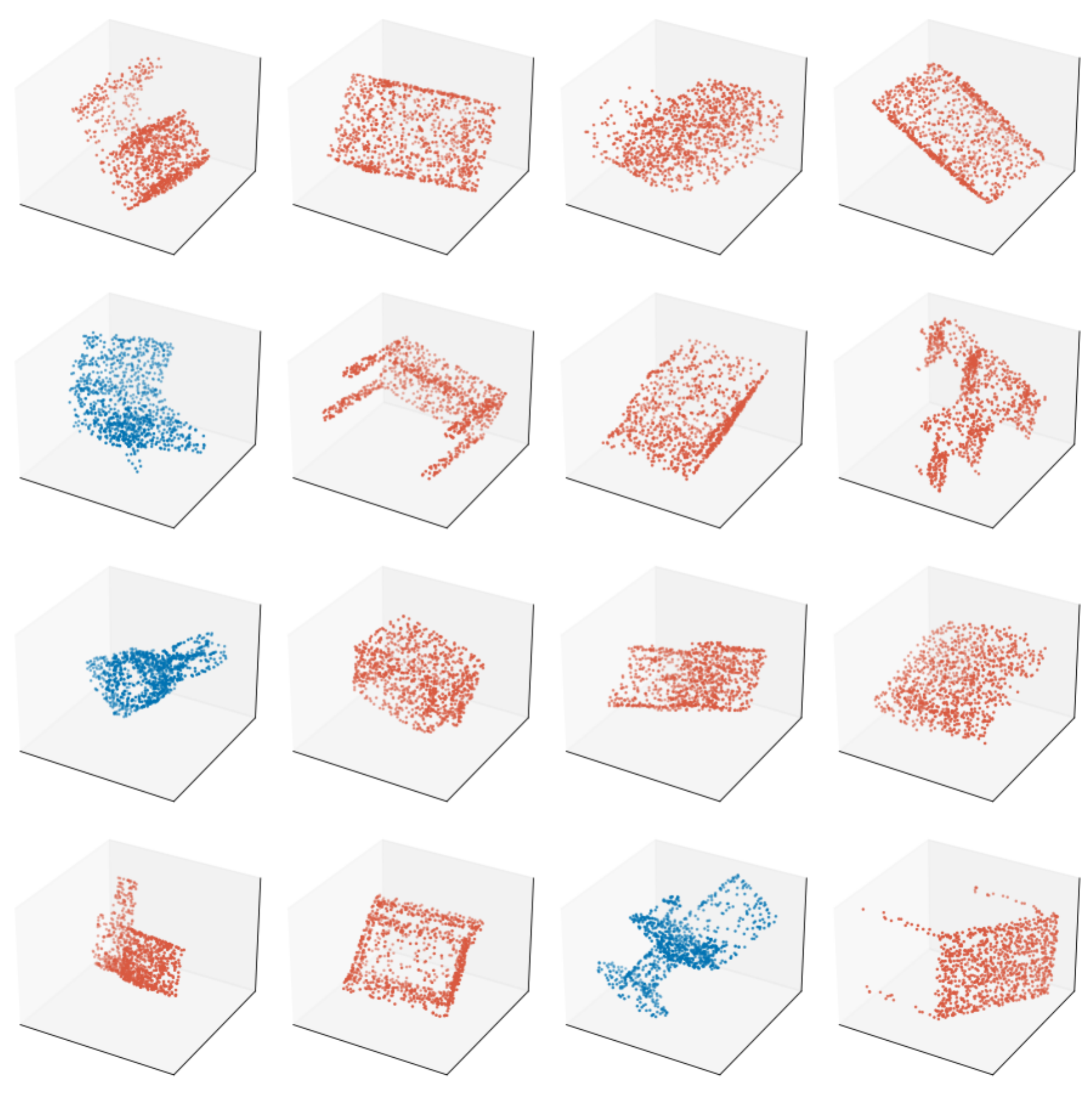}
    \caption{\centering ModelNet Transformed}
\end{subfigure}
\begin{subfigure}[t]{0.16\textwidth}
    \centering
    \includegraphics[width=\textwidth]{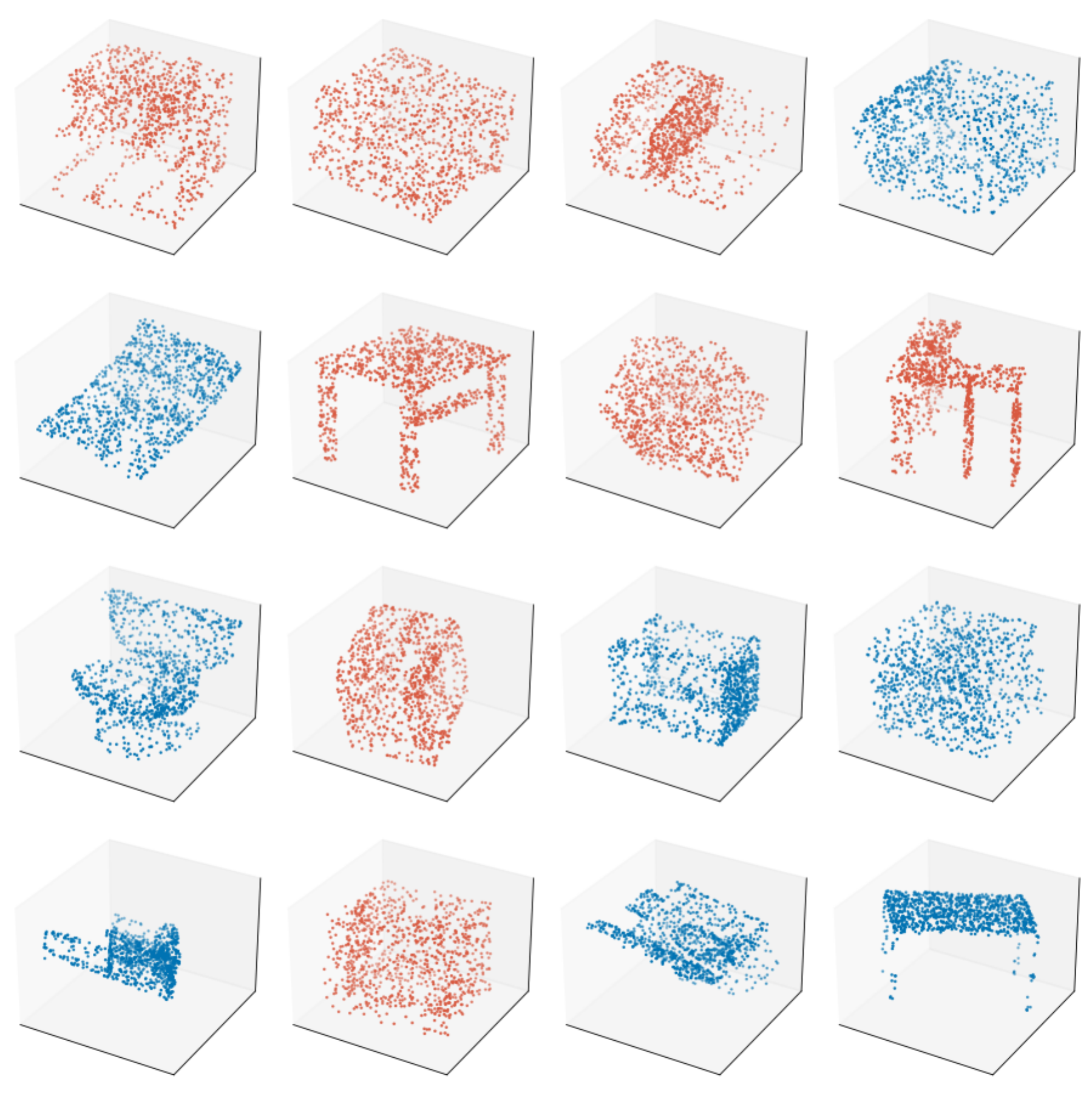}
    \caption{\centering ModelNet Canon}
\end{subfigure}
\begin{subfigure}[t]{0.16\textwidth}
    \centering
    \includegraphics[width=\textwidth]{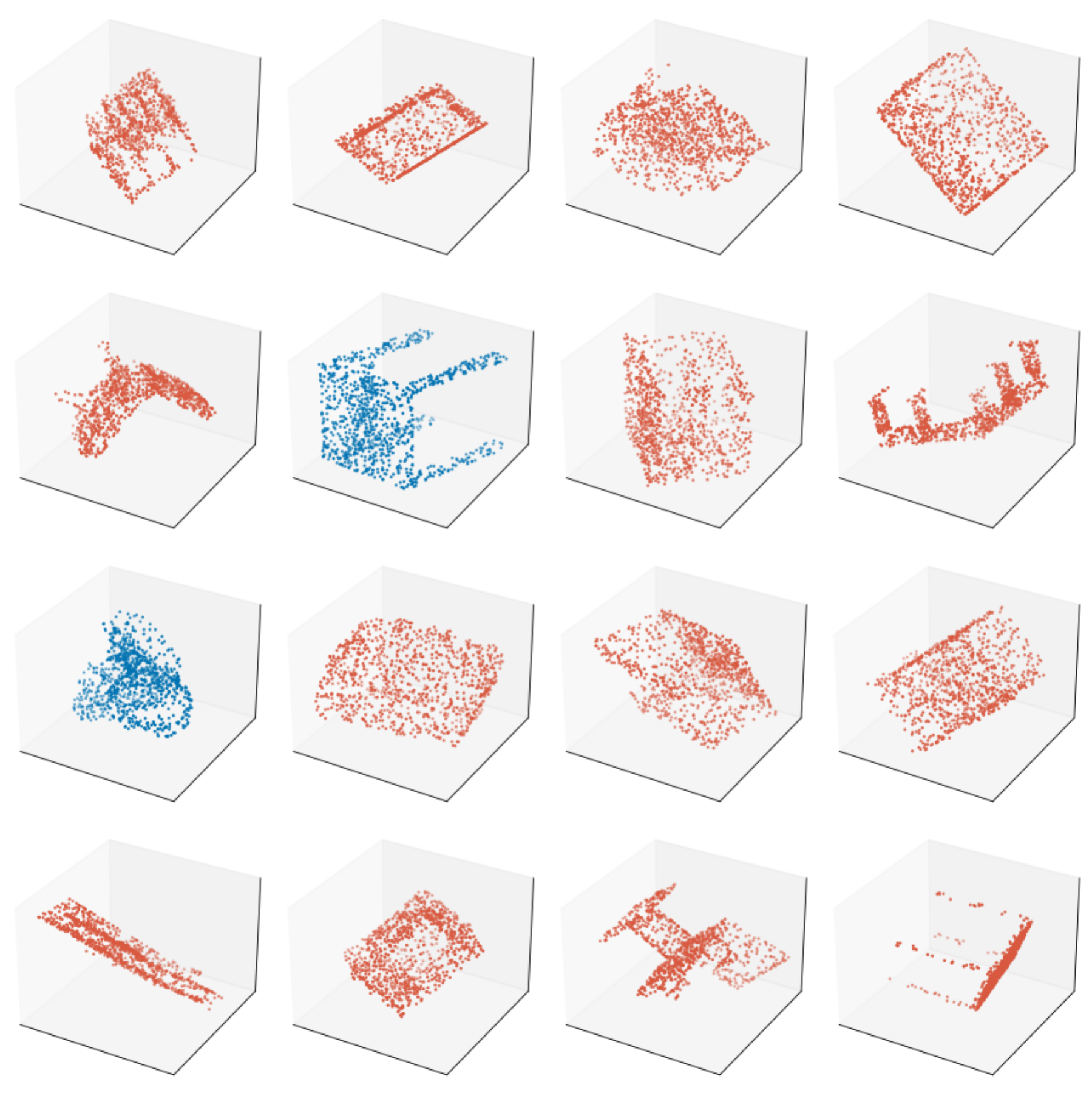}
    \caption{\centering ModelNet ITS}
\end{subfigure}

\caption[Examples of unsupervised canonicalization]
{Canonicalization results across datasets. Left: transformed inputs. Middle: canonicalization (Canon) using the best detector. Right: ITS results. Blue indicates correct classifications, red incorrect.}
\label{fig:unsupervised_all}
\end{figure}

As OOD scores exhibit varying performance depending on the dataset \citep{Yang2022}, we compare several detectors to examine whether this holds for canonicalization and whether a well-chosen score can improve results.
As the search strategy we use RS-LR as it worked well in previous experiments.
Hyperparameters of the detectors were optimized in a two-stage process (see \Cref{app:hyperparam}).
The results are shown in \Cref{fig:comparision_detectors_1}.
Distance-based methods like kNN and prototype-based methods outperform logit-based methods, with kNN-based methods performing best overall.
Laplace approximations and VIM also surpass logit-based methods but are beaten by distance-based approaches.
Activation reshaping methods show similar performance to logit-based scores. 
We show example results of the best unsupervised method from the main section using the same search procedure in \Cref{fig:unsupervised_all}. 
The results are qualitatively more aligned with the training data and more often classified correctly than with ITS~\cite{ITS}.
This shows that selecting an appropriate OOD score can improve both the accuracy and the alignment on transformed data.
A concern with logit- and feature-based OOD detection is that information loss in deeper layers may hinder canonicalization. 
In \Cref{app:supervised}, we show that training a learned score function on later-layer features reduces accuracy, suggesting that feature-based methods may be fundamentally limited compared to methods trained on the input data.

\subsection{How does Gating trade ID vs OOD performance?}
\label{sec:exp:gating}


\begin{figure}[t]
   \centering
   \begin{subfigure}{0.48\linewidth}
       \centering
       \includegraphics[width=\linewidth]{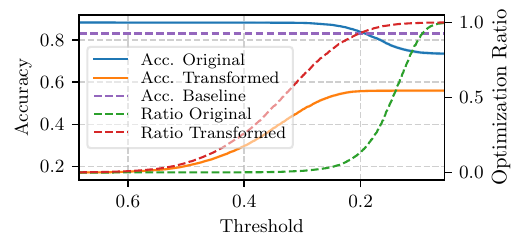}
   \end{subfigure}
   \begin{subfigure}{0.48\linewidth}
       \centering
       \includegraphics[width=\linewidth]{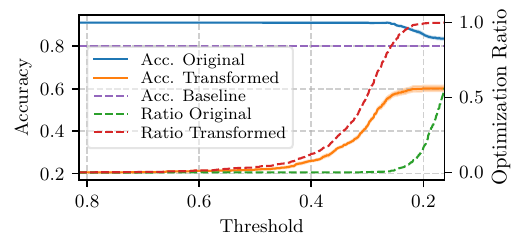}
   \end{subfigure}
   \caption{Trade-off between accuracy on original and transformed data when varying the OOD threshold for EMNIST (left) and ModelNet10 (right), showing gains on transformed data without much loss on original. The baseline is the accuracy of the augmented model on original data.}
   \label{fig:confidence_threshold_comparison}
\end{figure}

A working OOD detection setup should not degrade a model's performance on ID data~\citep{Yang2022}.
By thresholding the OOD score and only canonicalizing samples deemed OOD, we can control the trade-off between performance on transformed and untransformed data, accepting search results only when the score improves.
Due to a lack of datasets with realistic mixtures of transformations, we only report the threshold's effect on the standard and transformed test sets.
\Cref{fig:confidence_threshold_comparison} shows this trade-off for EMNIST and ModelNet10 using the best unsupervised OOD score per dataset.
This shows that most performance on untransformed data can be retained while significantly improving it on transformed data, though it remains below always canonicalizing since some transformed inputs are skipped, which reduces the computational overhead.
Other datasets show similar trends, see \Cref{app:selectiongate}.

\section{Conclusions}
Zero-shot test-time canonicalization can and should take advantage of the vast space of existing out-of-distribution (OOD) scores and search heuristics.
Selecting an appropriate OOD score significantly improves performance compared to previous work \citep{ITS}.
We found that random search with local refinement coupled with distance-based OOD scores performs best overall, while the single best score is dataset dependent.
Furthermore, existing canonicalization methods \citep{STN} tend to degrade in-distribution (ID) performance as all inputs are always transformed.
We introduce a gating mechanism that ensures that canonicalization is only executed when needed, significantly mitigating the negative effect on the ID performance.
The framework requires no retraining and treats the classifier as a black box, so it can be attached to any pretrained model to increase robustness, with the inference-time search cost being the main practical consideration.

\subsubsection*{Limitations and Future Work}
While this framework requires no retraining, the 60-sample inference budget may limit real-time applicability. 
Future work could explore more sample-efficient alternatives such as evolutionary methods or Bayesian optimization. 
Relying solely on deep features can be insufficient when spatial information is critical for canonicalization. Combining OOD scores with energy scores from foundation models~\cite{Singhal2025} is a promising direction. 
When transformations are known a priori and retraining is feasible, augmentation-based training can still be preferable, though test-time canonicalization offers a compelling path toward closing this gap without such assumptions.
Canonicalization might also be applied locally to parts or individual objects rather than the entire frame using part sampling methods~\cite{Saccader}.

\begin{credits}
\subsubsection{\discintname}
The authors have no competing interests to declare that are relevant to the content of this article.
\end{credits}

\section{Acknowledgments}
Dominik Lindner was funded by the IB (ZS/2026/08/204393).
Johann Schmidt was funded by the ZIM (KK5540302MS4).
Tom Siegl was cofunded by the BMFTR (01IS22077, 01ED2507) and DFG SFB 1713.

%
%
%
\bibliographystyle{splncs04}
\bibliography{main}
%
\newpage
\appendix




\section{Preliminaries Extended}
\label{app:preliminaries}

\subsection{Group Theory and Affine Transformations}
\label{app:grouptheory}
The affine transformations we consider are described by matrix Lie groups. 
We refer to works such as \cite{Hall2015} for a more detailed introduction to the topic.
An example is the interval $[0,2\pi)$, which is a Lie group under addition modulo $2\pi$.
These groups act on other sets via a group action, i.e. a map $G \times \mathcal{X} \to \mathcal{X}$ that transforms the data according to the group element.
For the example above, one group action is the rotation about the image center, where the group element is the rotation angle and the data is the image itself.
In general, the transformations of interest can be represented by $(n+1)\times(n+1)$ affine transformation matrices that act on the homogeneous coordinates of the data.
If the action is clear from context, we write $gx$ instead of explicitly specifying the group action.
We generally consider groups generated by reflections, rotations, shearing, scaling, and translations.
In 2D, we parameterize rotations using angles in $[0, 2\pi)$. For 3D data, we use the representation from  Zhou et al. \citep{Zhou2020} to avoid issues such as the double covering of $SO(3)$ by quaternions or discontinuities in Euler angles.

\subsection{Canonicalization}
\label{app:canonicalization}
Definition \ref{defn:Canonicalization} defines canonicalization by mapping to a representative member.
Kaba et al. \citep{Kaba2023} used a slightly more flexible definition describing how an equivariant function can be defined by mapping to group elements:
\begin{definition}[Equivariance via Canonicalization \cite{Kaba2023}]\label{defn:Canonicalization2}
    Given a function $c: \mathcal{X} \to G$ and a prediction function $f: \mathcal{X} \to \mathcal{Y}$, a function $\phi: \mathcal{X} \to \mathcal{Y}$ defined by
    \[
        \phi(x) = c(x) \cdot f(c(x)^{-1} \cdot x)
    \]
    is equivariant if $c$ is $G$-equivariant.
\end{definition}
Here, $c$ is called the canonicalization function, which maps to the group elements.
This definition was not used in the main part as it does not directly demonstrate that canonicalization can be interpreted as mapping to orbit representatives.
An orbit representative is obtained by applying $x \mapsto c(x)^{-1} \cdot x$.

For instance, take the Lie group of continuous rotations $G=SO(2)$ and an image of a circle.
Then, any rotation leaves the circle unchanged, $ g \cdot x = x, \forall g \in G$.
The stabilizer of $x$ is the entire group $G$, which is non-trivial. As such $x$ is symmetric.
In such cases, the $G$-equivariant $c$ will fail, as for any non-identity element $g \in G$, we have
\begin{equation}
    c(x) = c(g \cdot x) = g \cdot c(x).
\end{equation}
Right-multiplying by $c(x)^{-1}$ gives $e = g$, which contradicts $g \neq e$, where $e$ is the identity element of $G$.
Therefore, an equivariant canonicalization function cannot exist for these symmetric samples.
Kaba et al.~\citep{Kaba2023} show this for all groups with non-trivial stabilizers.
To address this, they introduced relaxed equivariance, a weaker form requiring equivariance only for asymmetric inputs.
\begin{definition}[Relaxed Equivariance \cite{Kaba2023}]
    A function $f: \mathcal{X} \to \mathcal{Y}$ is relaxed equivariant to a group if for all $g_1 \in G, x \in \mathcal{X}$ there exists a $g_2 \in g_1 G_x$, where
    \[
        G_x = \{ g \in G \mid g \cdot x = x \}, \quad
        g_1 G_x = \{ g_1 \cdot g \mid g \in G_x \},
    \]
    where $G_x$ is the stabilizer of $x$, such that $f(g_1 \cdot x) = g_2 \cdot f(x)$.
    \label{defn:relaxed_equivariance}
\end{definition}

When $G_x = \{e\}$ (asymmetric input), $g_2 = g_1$ exactly, and the definition reduces to standard equivariance.
In the invariant special case ($\mathcal{Y}$ carries the trivial representation), $g_2 \cdot f(x) = f(x)$ trivially and the condition collapses to invariance.
If the canonicalization function $c$ is relaxed equivariant, then the function $\phi$ from \Cref{defn:Canonicalization2} remains relaxed equivariant as well~\citep[Theorem A.2]{Kaba2023}. 
For asymmetric inputs ($G_x = \{e\}$), the coset collapses to the singleton $g_1 G_x = \{g_1\}$, so the relaxed condition 
reduces to the standard equivariance requirement $c(g \cdot x) = g \cdot c(x)$.
We can show that this results in invariance for canonicalization when defined via \Cref{defn:Canonicalization}.
According to this definition, we have a canonicalization function $h: \mathcal{X} \to \mathcal{X}$ with $h(x) = c(x)^{-1} \cdot x$.
We assume relaxed equivariance for $c: \mathcal{X} \to G$, where $G$ acts on itself by left multiplication, i.e.\ $g \cdot c(x) := g \cdot_G c(x)$.
We have
\begin{equation}
\begin{alignedat}{2}
    h(g_1 \cdot x) &= c(g_1 \cdot x)^{-1} \cdot (g_1 \cdot x)
       && \text{(def. of $h$)} \\
    &= (g_1 g' c(x))^{-1} \cdot g_1 \cdot x
       && \text{($g_2 = g_1 g',\ g' \in G_x$)} \\
    &= c(x)^{-1} (g')^{-1} g_1^{-1} g_1 \cdot x
       = c(x)^{-1} (g')^{-1} \cdot x \\
    &= c(x)^{-1} \cdot x = h(x)
       && \text{($g' \in G_x$)}
\end{alignedat}
\end{equation}
Hence, $h$ is invariant and maps to a representative (canonical form) in $\mathcal{X}$.

\subsection{Symmetry-Breaking}
\label{app:symmetry_breaking}
Two distinct sources of ambiguity must be separated here.
The first is the stabilizer: when $\mathrm{stab}(x) \neq \{e\}$, several group elements map $x$ to the same point, so the minimizer of $S$ is not unique.
This is harmless for our setting. 
All such elements yield the same canonical form, and an invariant classifier assigns them the same label.
Resolving this ambiguity, often called symmetry breaking, matters only for \emph{equivariant} downstream networks, where the canonical frame itself must transform consistently~\citep{Kaba2023}.
The second source is genuine: $S(g^{-1} \cdot x)$ may have multiple minima over $G$ that are \emph{not} related by any symmetry of $x$, so different initializations can return different canonical forms.
Principled remedies map to sets of representatives~\citep{Ma2024} or to Radon measures over group elements~\citep{Shumaylov2025} rather than committing to one.
But locating several well-separated minima is far more expensive than finding one, and even reliably finding a single near-optimal minimum for continuous groups is already non-trivial (see \Cref{sec:experiments}).
We therefore accept a single search result and rely on the gating mechanism (\Cref{sec:method}) to reject it when it fails to lower the OOD score.

\section{OOD Scoring Functions}
\label{app:ood}

\subsection{Categorization and Discussion}
\label{app:ood:discussion}

We focus on OOD scores computed from the logits or intermediate features of the base model (avoiding additional model training).
The scores fall into two groups: logit-based scores, which apply only to classifiers, and feature-based scores, which apply to any model.
We further divide the feature-based scores into rectification, prototype-based, and k-Nearest Neighbors (kNN)-based methods.
The methods differ in what they require: some are hyperparameter-free and need no validation set, while prototype- and kNN-based methods store untransformed training data.
We use the dominant category when methods come from several groups.

\subsubsection*{Logit-based methods}
Maximum Softmax Probability~\citep{HendrycksOOD2022} uses the maximum softmax probability as an inlier score, requiring no hyperparameters or training data.
The maximum logit (MaxLogit) is a similar alternative~\citep{Hendrycks2022}.
Advanced methods incorporate all outputs, such as the energy of the logits~\citep{Liu2020}, calculated as
$E(x) = -T \cdot \log\left(\sum_{i=1}^n e^{f_i(x)/T}\right)$, where $f_i(x)$ is the output of the model for input $x$ and
$T$ is an optional temperature parameter, typically set to 1 without tuning.
Similarly, the Shannon entropy of the softmax outputs can be used~\citep{Liu2023} as an OOD score.
Variants like GEN~\citep{Liu2023} incorporate temperature and a parameter limiting the number of contributing values, which requires hyperparameter tuning.
OpenMax~\citep{Bendale2016} fits a Weibull distribution to the logits to estimate OOD probability, requiring in-distribution data.
VIM~\citep{wangViMOutOfDistributionVirtuallogit2022}
modifies the energy score by incorporating feature-space information from the features before the final linear layer via a principal-subspace projection.
A residual is computed from the projection onto the orthogonal complement of this subspace.
A scaled version of this residual is added to the energy score.
Bayesian uncertainty estimation is another category.
While Bayesian networks require model changes, Monte Carlo Dropout~\citep{Gal2016} uses inference-time dropout to obtain uncertainty estimates.
Calculating this for the entire model is expensive, but applying dropout only to the final layers incurs a smaller overhead.
Dropout, combined with kernel-based smoothing, was previously used for canonicalization~\citep{ITS}.
Many models used here do not use dropout, so it was not included in our comparison.
Laplace Approximation~\citep{MacKay1992,Daxberger2021} is a Bayesian approximation not requiring model modification. Weight distributions are estimated by approximating the posterior as a Gaussian centered at the MAP estimate, with a regularization term tuned on a validation set.
Samples are obtained by sampling weights and passing the input through the modified model.
We use last-layer Laplace for efficiency~\citep{Kristiadi2020}.
The resulting output of the model is then processed via the entropy or mutual information (MI) criterion from TorchUncertainty~\citep{TorchUncertainty}.

\subsubsection*{Rectification-based methods}
These methods modify features during the forward pass, assuming OOD samples show different activation patterns, which can be amplified.
ReAct~\citep{ReAct2021} truncates activations above a threshold~\citep{Sun2022}.
This threshold is a hyperparameter, set to a percentile of activations on in-distribution data.
The modified features are passed through the model, and another OOD score (energy or entropy in our case) is applied.
DICE~\citep{DICE} ranks weights in the final layer by their contribution and disables non-contributing ones, requiring in-distribution data for assessing the contribution.
A validation set selects how many weights to disable.
ASH~\citep{Djurisic2023} prunes a percentile of activations during the forward pass, modifying others based on different modes.
Unlike ReAct, ASH only needs the percentile to prune, not a threshold.
This method requires hyperparameter tuning but no in-distribution data.

\subsubsection*{Prototype-based methods}
These methods calculate distances to class representatives, generally requiring the selection of a layer for feature extraction.
Simplified Hopfield Energy (SHE)~\citep{Zhang2023} stores the average feature per class and calculates the Hopfield energy as the dot product between the feature and the class average belonging to the predicted class.
Mahalanobis distance (MD)~\citep{Lee2018} computes the distance from a sample to each class mean, where distances are calculated using the inverse feature covariance matrix.
The minimum distance is used as the score.
The original work also used adversarial perturbations to increase the distance of OOD samples.
This is not applied due to the overhead of computing it when searching.
Relative Mahalanobis Distance (RMD)~\citep{Ren2021} modifies the Mahalanobis Distance by calculating the distance to the global mean and subtracting it from the class mean distance, improving performance on many datasets.
We also include a method called Prototype (Proto) that uses Euclidean, cosine, or a mixture of both distances to class means.
The mixture is $d = (1 - \alpha) \cdot d_{\text{euclidean}} + \alpha \cdot d_{\text{cosine}}$ where $\alpha$ is a hyperparameter. 
Additionally, we use a binary hyperparameter to determine whether to use squared Euclidean distance.
Similar to SHE, we consider the distance to the predicted class instead of the closest, which we call Per-class Prototype (PC-Proto).

\subsubsection*{kNN-based methods}
These methods~\citep{Sun2022} store all features of in-distribution data. The average distance to the $k$ nearest neighbors is calculated. 
The original work used Euclidean distance on normalized features. 
We use both Euclidean and cosine distance, equivalent to using normalized features with Euclidean distance.
The hyperparameters are the number of neighbors and the distance metric.
We consider variants using all classes (kNN), only the predicted class (PC-kNN), and a kNN Mix variant mixing Euclidean and cosine distance, similar to the prototype method.
Another variant is the Trust Score~\citep{Jiang2018}, which calculates the ratio of the distance to the nearest neighbor not of the predicted class divided by the distance to the nearest neighbor of the predicted class.
A limitation of kNN methods is the memory requirement for storing all features. 
We use float16 precision to reduce memory usage.

\subsection{Runtime}
\label{app:ood:runtime}

\begin{table}[t]
\caption{Relative inference time compared to Energy baseline (Energy = 1.00). Bottom row shows absolute time in ms for the baseline(per sample).}
\centering
\label{tab:speed_comparison}
\begin{tabular}{lcccc}
\toprule
 & MNIST & EMNIST & TU Berlin & ModelNet10 \\
\midrule
ASH & 1.15 & 0.99 & 1.03 & 0.92 \\
DICE & 1.12 & 1.08 & 1.02 & 1.02 \\
Energy & 1.00 & 1.00 & 1.00 & 1.00 \\
Entropy & 1.00 & 0.99 & 0.99 & 0.90 \\
GEN & 1.00 & 1.01 & 1.01 & 0.92 \\
KNN & 1.26 & 1.21 & 1.06 & 1.04 \\
KNN Mix & 1.50 & 1.81 & 1.08 & 1.11 \\
Lapl. Entropy & 1.52 & 1.55 & 1.51 & 1.39 \\
Lapl. MI & 1.63 & 1.60 & 1.46 & 1.53 \\
Lapl. Weighted & 1.71 & 1.61 & 1.48 & 1.48 \\
MD & 1.05 & 1.40 & 1.03 & 1.05 \\
MSP & 0.99 & 0.98 & 0.99 & 0.90 \\
MaxLogit & 1.00 & 1.01 & 0.99 & 0.93 \\
OpenMax & 1.52 & 1.40 & 1.18 & 1.20 \\
PC-Proto & 1.12 & 1.12 & 1.06 & 1.07 \\
PC-kNN & 1.32 & 1.32 & 1.07 & 1.08 \\
PC-kNN Mix & 1.62 & 1.88 & 1.11 & 1.13 \\
Proto & 1.04 & 1.00 & 1.04 & 1.07 \\
RMD & 1.06 & 1.18 & 1.01 & 1.06 \\
ReAct & 1.00 & 1.00 & 1.01 & 0.95 \\
SHE & 1.09 & 1.04 & 0.96 & 1.00 \\
Trust Score & 1.88 & 2.17 & 1.03 & 1.03 \\
VIM & 1.02 & 0.99 & 1.04 & 1.02 \\
\midrule
Baseline (ms) & 0.89 & 0.90 & 5.82 & 35.84 \\
\bottomrule
\end{tabular}
\end{table}

\begin{table}[t]
\caption{Profiler summary. Sz: size of parameters and buffers in MB (including the base model); VR: peak increase in VRAM during the forward pass in MB; GF: GFLOPs per batch.}
\label{tab:profiler_summary}
\begin{tabular}{l*{12}{wr{0.72cm}}}
\toprule
Dataset & \multicolumn{3}{c}{MNIST} & \multicolumn{3}{c}{EMNIST} & \multicolumn{3}{c}{TU Berlin} & \multicolumn{3}{c}{ModelNet10} \\
 & Sz & VR & GF & Sz & VR & GF & Sz & VR & GF & Sz & VR & GF \\
Component &  &  &  &  &  &  &  &  &  &  &  &  \\
\midrule
ASH & 18.7 & 94.3 & 27.8 & 18.8 & 94.3 & 27.8 & 9.6 & 226.2 & - & 5.6 & 289.4 & 23.7 \\
DICE & 18.7 & 94.3 & 27.8 & 18.8 & 94.3 & 27.8 & 9.6 & 225.2 & - & 5.6 & 287.8 & 23.7 \\
Energy & 18.7 & 94.3 & 27.8 & 18.8 & 94.3 & 27.8 & 9.6 & 226.2 & - & 5.6 & 289.4 & 23.7 \\
Entropy & 18.7 & 94.3 & 27.8 & 18.8 & 94.3 & 27.8 & 9.6 & 226.2 & - & 5.6 & 289.4 & 23.7 \\
GEN & 18.7 & 94.3 & 27.8 & 18.8 & 94.3 & 27.8 & 9.6 & 226.2 & - & 5.6 & 289.4 & 23.7 \\
KNN & 229.7 & 134.2 & 45.3 & 217.1 & 149.8 & 41.1 & 40.8 & 225.2 & - & 12.6 & 287.2 & 23.7 \\
KNN Mix & 229.7 & 120.4 & 43.3 & 208.3 & 214.3 & 41.3 & 40.6 & 225.2 & - & 19.4 & 287.6 & 23.8 \\
Lapl. Entropy & 18.7 & 95.7 & 28.6 & 18.8 & 100.2 & 31.7 & 9.6 & 265.3 & - & 5.6 & 288.8 & 23.8 \\
Lapl. MI & 18.7 & 95.7 & 28.6 & 18.8 & 100.2 & 31.7 & 9.6 & 266.2 & - & 5.6 & 290.0 & 23.8 \\
Lapl. Weighted & 18.7 & 95.7 & 28.6 & 18.8 & 100.2 & 31.7 & 9.6 & 266.2 & - & 5.6 & 288.8 & 23.8 \\
MD & 21.2 & 94.3 & 29.2 & 23.6 & 106.8 & 39.6 & 12.7 & 226.2 & - & 10.9 & 289.1 & 24.1 \\
MSP & 18.7 & 94.3 & 27.8 & 18.8 & 94.3 & 27.8 & 9.6 & 226.2 & - & 5.6 & 289.4 & 23.7 \\
Main Model & 18.7 & 94.3 & 27.8 & 18.8 & 94.3 & 27.8 & 9.6 & 225.2 & - & 5.6 & 287.8 & 23.7 \\
MaxLogit & 18.7 & 94.3 & 27.8 & 18.8 & 94.3 & 27.8 & 9.6 & 226.2 & - & 5.6 & 289.4 & 23.7 \\
OpenMax & 18.7 & 94.3 & 27.8 & 18.8 & 94.3 & 27.8 & 9.8 & 226.2 & - & 5.6 & 289.4 & 23.7 \\
PC-Proto & 18.8 & 94.3 & 29.0 & 18.9 & 94.3 & 28.4 & 10.0 & 226.2 & - & 5.6 & 289.4 & 23.7 \\
PC-kNN & 229.9 & 134.2 & 43.6 & 217.5 & 150.1 & 42.0 & 40.9 & 225.2 & - & 12.6 & 287.3 & 23.7 \\
PC-kNN Mix & 229.9 & 120.4 & 43.7 & 208.7 & 214.3 & 41.3 & 40.9 & 225.2 & - & 19.4 & 288.4 & 23.8 \\
Proto & 18.8 & 94.3 & 29.0 & 18.9 & 94.3 & 27.8 & 10.0 & 226.2 & - & 5.6 & 289.4 & 23.7 \\
RMD & 23.2 & 94.8 & 29.3 & 22.2 & 94.3 & 31.9 & 12.1 & 226.2 & - & 8.2 & 289.4 & 23.8 \\
ReAct & 18.7 & 94.3 & 27.8 & 18.8 & 94.3 & 27.8 & 9.6 & 226.2 & - & 5.6 & 289.4 & 23.7 \\
SHE & 18.7 & 94.3 & 29.0 & 18.8 & 94.3 & 28.4 & 9.6 & 225.2 & - & 5.6 & 287.8 & 23.7 \\
Trust Score & 18.7 & 196.4 & 40.7 & 18.8 & 275.7 & 40.7 & 9.6 & 225.2 & - & 5.6 & 289.4 & 23.7 \\
VIM & 18.7 & 94.3 & 27.9 & 18.8 & 94.3 & 27.8 & 9.6 & 225.2 & - & 5.6 & 287.8 & 23.7 \\
\bottomrule
\end{tabular}

\end{table}

While many OOD scores only have negligible overhead, especially distance-based methods can incur significant VRAM and computational overhead. 
\Cref{tab:speed_comparison} shows the execution time when performing the search using RS-LR on a batch of samples (budget 60). 
Results were calculated over 50 batches for each detector and averaged.
Generally, kNN-based methods have a significant overhead when used together with smaller models where the baseline is very fast.
Using mixed distances can increase the cost further, as our implementation is not optimized and calls cosine and Euclidean distance individually (when the mixing value is between 0 and 1).
kNN-based methods generally use 16-bit precision to reduce VRAM usage and computation time. 
We did not implement it for the Trust Score, explaining the larger overhead.
The per-class version uses masking after calculating the full distance matrix. A custom operation that calculates the distance only to the predicted class should be able to reduce the overhead.
Since kNN scales with the number of stored samples, subsampling can be used to reduce computational overhead on larger datasets.

We tested the PC-kNN detector on SI-Score, which used a subset of ImageNet, and measured an overhead of less than 5\% compared to the energy of the logits.
Alternatively, approximate nearest neighbor methods like FAISS~\citep{Douze2024} can likely reduce computation.
For our experiments on MNIST, we did not observe a speedup using FAISS, hence we relied on brute force search.
We compared the computation footprints in \Cref{tab:profiler_summary} (batch size: 128 for MNIST, EMNIST, and TU Berlin, 16 for ModelNet10).
Note that for the BiLSTM, the GFLOPs were not measured correctly and have therefore been omitted.
For the last-layer Laplace approximation, we looped over the last linear layer and aggregated the results. This explains the overhead in measured runtime compared to the FLOPs.
It can be further optimized by calculating the results in parallel.
MD and RMD are the only algorithms in our implementation that compute layer-wise distances using features from multiple layers and combine these distances linearly, following the original work.



\section{Search Algorithms}
\label{app:search}

\subsection{Categorization and Discussion}
\label{app:search:discussion}

For canonicalization of continuous groups, global optimization is needed, as the definition requires global minima. Otherwise, the global minimum may not be reliably recovered from every orbit member. 
The energy landscapes are generally non-convex, see \Cref{app:energy_landscapes} for examples.
Different search algorithms differ in how they handle non-continuous parameterizations (like reflections, which are not connected) and unbounded search spaces. Most algorithms are initialized in a region around the origin. Whether they can leave this region and explore the full unbounded search space depends on the algorithm.

\subsubsection*{Multi-start gradient descent}
The algorithm runs multiple instances of gradient descent in parallel.
Multi-start gradient descent handles disconnected group elements by starting in different connected components but cannot move between them.
It can handle unbounded spaces only if the function is convex, which is rare.
We use Adam \citep{Adam} to handle varying gradient scales.
The hyperparameters are the learning rate, the number of random starts, and the step count.

\subsubsection*{Simulated Annealing}
Simulated Annealing is a stochastic global optimization algorithm.
Using it for canonicalization was first proposed by Schmidt \& Stober \cite{ITS} but not tested. We use a variant that samples around the neighborhood of a point per iteration. Better values are always accepted, while worse values are accepted depending on the temperature, which is lowered over time.
It can handle connected components by moving between them during optimization.
Non-compact search spaces are handled by allowing moves outside the initial regions.
While versatile, it requires many iterations, resulting in high latency.
To reduce latency, we use a version where we run multiple instances of simulated annealing in parallel.
The hyperparameters are the initial temperature, cooling rate, neighborhood size, number of parallel runs, and step count.
We use a rectangular neighborhood with a size relative to the initial domain, so we only have one parameter for neighborhood size.

\subsubsection*{Random search}
This method randomly samples points in the search space and returns the best one. Our implementation uses Sobol sequences to better cover the search space.
Random search handles disconnected components by sampling in all components.
However, non-compact search spaces are difficult to handle.
A distribution like a Gaussian could be used, but few samples would fall far from the center. 
As such, we assume only the bounded case and sample in a fixed region.
This search has only the number of samples as a hyperparameter.

\subsubsection*{Random search with local refinement}
Random search with local refinement refers to a two-step process: first sampling multiple points randomly and then refining the best points using local optimization. We use gradient descent with the Adam optimizer\citep{Adam}.
It covers disconnected components similarly to random search.
However, the local optimization stays in the same connected component.
In general, while one can move slightly outside the initial domain, the low step counts make this unsuitable as a method when domain restrictions are not assumed.
Hyperparameters include the number of initial samples, the number of points to refine, the learning rate, and the step count for the local optimization.

\subsubsection*{Coordinate Descent and Variants}
The use of coordinate-descent-like algorithms for canonicalization, though not being global optimizers, was first introduced by the Inverse Transformation Search (ITS) algorithm \citep{ITS}.
Coordinate descent works by iterating over each dimension and using a per-dimension optimizer on that dimension. 
Unlike typical coordinate descent, we sample points across the whole domain of each coordinate and select the best, mirroring the ITS implementation.
ITS starts multiple searches by selecting the top $k$ points in the first dimension and branching to refine each individually.
We investigate variants that differ from the ITS algorithm in
the original work.
Coordinate Descent only refines a single top point.
For the per-dimension search, we randomly sample points across the entire domain while also comparing against the previous value.
Unlike ITS, our coordinate descent implementation includes a parameter for the number of cycles, enabling multiple passes over all dimensions.
As an alternative sampling strategy, we consider equally spaced points but randomly offset the first point to allow multiple runs (lattice sampling).
Similar to ITS, which assigns more importance to the first dimension, we consider Weighted Coordinate Descent (WCD), which increases the sampling density in the first dimension relative to others.
We call the variants CD-S when performing one pass,
CD for unweighted coordinate descent, WCD for the weighted variant, and WCD-L for the weighted variant with lattice sampling.
All variants can handle disconnected components if a specific dimension corresponds to switching between them (e.g., reflection). In our implementation, they always assume a fixed domain and cannot handle unbounded ones.
Hyperparameters include: number of cycles (1 for CD-S), samples per dimension, and the weighting factor for the first dimension (for WCD and WCD-L).

\subsubsection*{Decomposition Assumption}
The sampling of initial points is typically done by assuming a decomposition into subgroups. We typically use rotation, shearing, followed by scaling.
Coordinate descent-based algorithms, however, refine each subgroup iteratively. Like ITS, this makes them very sensitive to the order of the transformations.
The other methods can also be affected, as the decomposition is used for sampling, for the neighborhood decomposition, and it defines the parameterization that is used for gradient descent.
We generally use the same generation process for the transformations during the search as was used to generate the transformed datasets (see \Cref{app:restricted_search_space}).

\subsection{Runtime and Complexities}
\label{app:search:runtime}

\begin{table}[t]
\centering
\caption{Mean processing time per batch (seconds) across different datasets and algorithms. PSO is particle swarm optimization.} 
\label{tab:benchmark_results}
\setlength{\tabcolsep}{5pt}
\begin{tabular}{l*{10}{c}}
\toprule
& \rotatebox{90}{CD}
& \rotatebox{90}{CD-S}
& \rotatebox{90}{GD}
& \rotatebox{90}{ITS}
& \rotatebox{90}{SA}
& \rotatebox{90}{PSO}
& \rotatebox{90}{RS}
& \rotatebox{90}{RS-LR}
& \rotatebox{90}{WCD}
& \rotatebox{90}{WCD-L} \\
\midrule
MNIST      & 0.13 & 0.12 & 0.13 & 0.14 & 0.18 & 0.11 & 0.11 & 0.12 & 0.15 & 0.15 \\
EMNIST     & 0.13 & 0.12 & 0.13 & 0.14 & 0.15 & 0.12 & 0.11 & 0.12 & 0.16 & 0.14 \\
ModelNet10 & 0.57 & 0.57 & 0.49 & 0.61 & 0.56 & 0.54 & 0.54 & 0.55 & 0.58 & 0.64 \\
TU Berlin  & 0.75 & 0.80 & ---  & 0.85 & 0.89 & 0.82 & 0.65 & ---  & 0.79 & 0.74 \\
\bottomrule
\end{tabular}
\end{table}

While the budget was limited to the same amount for all search methods,
the execution time can still differ due to some methods benefitting more from parallelization or having higher algorithmic overhead. \Cref{tab:benchmark_results} shows the time it takes to optimize a single batch (batch size: 128 for MNIST, EMNIST, and TU Berlin, 16 for ModelNet10.) over different datasets.
The reported time is averaged over the three score functions used in the search experiment (\Cref{sec:exp:search}): logit energy, PC-kNN, and the learned energy model.


\section{Implementation Details}
\label{app:implementation}

\subsection{Model Architectures and Training Details}
\label{app:architecture_training}

For the MNIST variant, we use a small residual network with 4 residual blocks and 64, 128, 256, and 512 channels.
For TU Berlin, we use a bidirectional LSTM~\citep{LSTM} with 2 layers and hidden size 256 followed by 2 fully connected layers with 512 units.
For ModelNet10, we use PointNet++~\citep{PointNetPP} with the implementation from the PyTorch Geometric library~\citep{Fey2019}.
The models were trained using Adam~\citep{Adam} with default hyperparameters for 100 epochs for MNIST/EMNIST, 50 epochs for TU Berlin, and 200 epochs for ModelNet10.
The best model on the validation set, saved per epoch, was used for all experiments.
For PointNet++, we disable random sampling during evaluation so that the scores are deterministic.
A single base network is trained per dataset using cross-entropy loss without data augmentation and used for all experiments.
For feature-based OOD detection, we use random projection and global pooling.
All datasets are divided into training, validation, and test sets:
MNIST/EMNIST: 90\% train, 10\% validation; separate test sets.
ModelNet10: 90/10 split of the training set; predefined test set.
TU Berlin: 80\% train, 10\% validation, 10\% test.
For SI-Score, we used a subset of ImageNet data (95,476 images total) by randomly sampling images from the ImageNet-1k training set until we had at least 50 images per class but kept at most 100 images per class.
90\% of this data was used to calculate embeddings and 10\% was kept for hyperparameter tuning.
The SI-Score dataset is only used as a transformed test set.
We used a batch size of 128 except for ModelNet10, where a batch size of 16 was used.

\subsection{Feature Extraction for OOD}
\label{app:feature_extraction}
Many OOD scores rely on features from earlier layers.
For the ResNet on MNIST, we use the input of the last activation of each block as well as the input of the final fully connected layer,
as this performed well for ASH~\citep{Djurisic2023}.
For the ResNet on SI-Score we use the output of the last activation at each block and the input of the final fully connected layer.
For PointNet, we consider the features after point pooling and from the subsequent fully connected layers.
Similarly, for the BiLSTM, we use the features of the fully connected layers after the LSTM.
Due to the large size, using complete features for image data is not feasible.
The Mahalanobis Distance averages over spatial dimensions \cite{Lee2018}, which can remove useful spatial information.
We employ spatial pooling to a spatial size of $3\times 3$ followed by random projection to a fixed size.
This shows improvement for SI-Score over global average pooling (\Cref{app:pooling}).
For small feature maps ($<$20000 elements), we consider random projection on the entire feature map.
The final sizes selected were 1024 for MNIST, 512 for EMNIST, and 4096 for SI-Score.
The selected layer, the reduction method, and whether to use all features or only those from correctly classified samples are all hyperparameters.

\subsection{Hyperparameter Tuning}
\label{app:hyperparam}
Random Search with Local Refinement with 46 initial samples and 2 points for local refinement with 3 steps was used when optimizing the hyperparameters of the OOD scores.
We first tested 40 hyperparameter configurations (trials) on three 1/6-sized subsets of the full validation data, then ran a 15-trial optimization on the full validation set using Optuna’s MedianPruner~\citep{Akiba2019}, seeded by enqueuing the three best candidates from each subset search. 
The validation data was split from the training data. The same transformations as the test data were applied to the validation data.
We show in \Cref{app:unsupervised_ood} that this fails: hyperparameters that are good for detection are not necessarily good for canonicalization, so we tune directly on the search objective.

\section{Further Experiments}
\label{app:experiments}

\subsection{Choice of Pooling}
\label{app:pooling}

This experiment compares pooling for feature-based OOD detection on the SI-Score rotation dataset.
The SI-Score rotation dataset with a PC-kNN detector (3 neighbors, mixed distance, weight 0.5) was used. 
A ResNet-50 pretrained on ImageNet is the classifier.
The output of the last block (before final global average pooling) was used as features.
Only features from correctly classified data points were used.
We compare global average pooling (over all spatial dimensions) against adaptively averaging to $3\times3$ and then randomly projecting to 4096 dimensions.
Global pooling results in a 2048-dimensional feature vector.
For optimization, we use sampled equidistant points as we have a single dimension.
Adaptive average pooling to $3\times3$ followed by random projection to 4096 dimensions reaches $60.7 \pm 0.1\%$ top-1 accuracy, significantly above the $57.8 \pm 0.2\%$ of global average pooling, justifying its use as the feature reducer in this work.

\subsection{Using OOD detection for canonicalization}
\label{app:unsupervised_ood}
This section examines the central assumption of our method: that minimizing an OOD score recovers a correct canonical form. 
Even for asymmetric inputs there is no guarantee that the minimal score is unique across the orbit, which orbit canonicalization requires~\cite{Kaba2023}.
Practically, we can only approximate the minima using optimization methods, making this less of a concern.
Unlike OOD detection, which thresholds activations to identify samples, canonicalization requires minimizing the score across an orbit.
Consequently, metrics effective for OOD detection may not find optimal solutions for canonicalization.
Hyperparameters of OOD scores that work well for OOD detection may not perform well for the task of canonicalization.
To evaluate this, we use both the transformed validation data, which we treat as OOD, and the original validation data, that we treat as in-distribution.
We evaluate the AUROC and the false positive rate at 95\% true positive rate for hyperparameter optimization of OOD scores for canonicalization.
Additionally, we calculate the accuracy when comparing the OOD score of a transformed image to the original image, counting it as correct if the original image has a lower OOD score, which does not require thresholding.

\begin{figure}[t]
    \centering
    \includegraphics[width=0.8\textwidth]{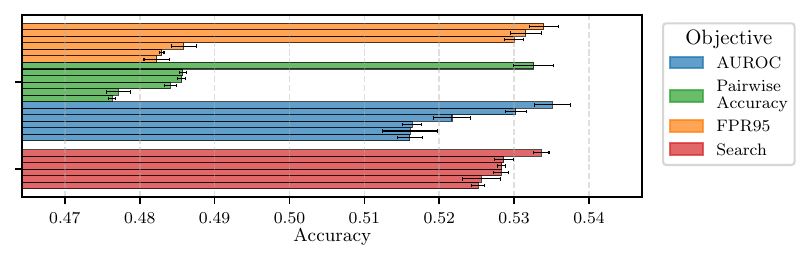}
    \caption{Comparison of different metrics for hyperparameter optimization of OOD scores when applied to canonicalization on EMNIST.}
    \label{fig:emnist_hyperparam}
\end{figure}

Figure \ref{fig:emnist_hyperparam} shows the results of 6 hyperparameter optimization runs (30 evaluations each).
These show that OOD metrics do not find optimal hyperparameters for canonicalization.
The search can also find hyperparameters that vary slightly in accuracy over the runs, likely due to the stochastic nature of both the hyperparameter search and the optimization methods.
From a practical perspective, the computation required for the hyperparameter tuning is a significant challenge, as the search is computationally expensive.

\subsection{Supervised Learning Canonicalization Functions}
\label{app:supervised}
To our knowledge, there are no supervised canonicalization functions proposed in the literature that operate on continuous groups.
Previous works(e. g. \citep{Panigrahi2024})  were limited to discrete groups or required specific architectures like an Input Convex Neural Network (ICNN)~\citep{Amos2017}.
Similar to LieLAC\citep{Shumaylov2025}, we learn the energy as a classifier but do not require a specific architecture.
We refer to this approach as learned energy model.
For the learned energy model we use the base model’s architecture but replace the final layer by a linear one with a single output.

\begin{figure}[t]
    \centering
    \includegraphics[width=0.6\textwidth]{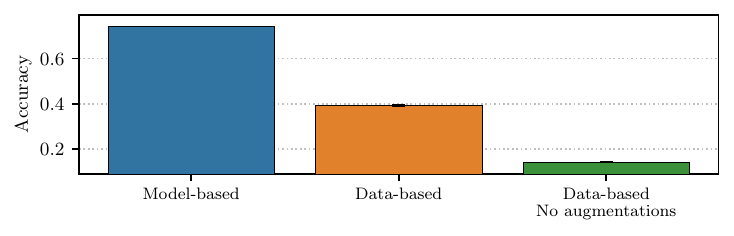}
    \caption{Comparison of learned energy models trained using either base-model predictions (model-based) or untransformed and transformed data (data-based) to determine labels on the EMNIST dataset. A variant without augmentations is included for comparison.
    }
    \label{fig:dec_strat_vs_no_dec_strat}
\end{figure}

A simple objective would be binary cross-entropy to distinguish between original and transformed data.
However, this can lead to only narrow ranges around the data being assigned a high score, which is suboptimal as narrow ranges make finding the minima more difficult.
To prevent such collapse, we change the objective to predicting whether the base model predicts the correct class.
We concatenate the original and transformed data in a batch and use the base model to assign labels.
The disadvantage is that this requires running the base model to get predictions during training.
Canonicalization needs a unique minimum up to symmetries.
We rely on the model's limited capacity to avoid assigning low energy everywhere.
For image data, due to its discrete nature, there are easily exploitable artifacts from interpolation and extrapolation.
For interpolation artifacts, we apply small affine transformations (one-pixel translation, three-degree rotation, small scale changes) to the original data to add similar artifacts.
Additionally, we apply random blur and unsharp masking to both original and transformed data to further prevent the model from relying on artifacts.
This only targets interpolation artifacts.
Extrapolation artifacts are more difficult to handle and require more complex setups like masking the foreground object and inserting it into a new background, as used in SI-Score.
\Cref{fig:dec_strat_vs_no_dec_strat} shows the canonicalization result on EMNIST when using random search with local refinement with the same hyperparameters as in \Cref{app:ood} combined with the learned energy.
The models were trained for half the epochs of the base model with the same settings as the base model.
Using base classifier predictions and augmentations helps the model learn a more useful energy function for canonicalization.

\paragraph{Details for learned energy in main comparison}
An advantage of learned models is that the size of the model can be chosen freely. We found that reducing the number of channels by a factor of 4 works well for the datasets when increasing
the evaluations during the search by the same factor.
Note that for LSTM we did not decrease the channel size of the MLP as it is not the limiting factor, and for PointNet we do not reduce the channels on the layer when averaging over points to avoid introducing an information bottleneck.
This also reduces computation cost for ResNet-based models and LSTM by a factor equal to or larger than 4.
For PointNet++, the reduction was smaller, around 3×, so we only increased the budget by that amount.
We use random search with 195 initial samples and 5 points further optimized using 4 local steps for most models.
For PointNet++ we use 153 initial samples and 3 refined using 4 local steps.
For the LSTM we use 240 random samples due to LSTMs requiring unrolling for gradient calculation. 
This is the configuration that is used for the comparison in \Cref{sec:exp:benchmark}.

\begin{figure}[t]
    \centering

    \begin{subfigure}[b]{0.4\textwidth}
        \centering
        \includegraphics[width=\textwidth]{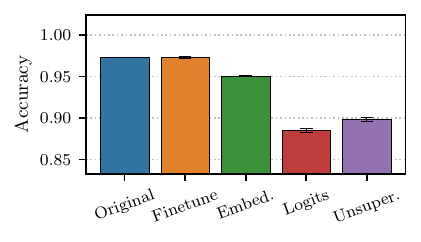}
        \caption{MNIST}
        \label{fig:accuracy_comparison_mnist}
    \end{subfigure}
        \begin{subfigure}[b]{0.4\textwidth}
        \centering
        \includegraphics[width=\textwidth]{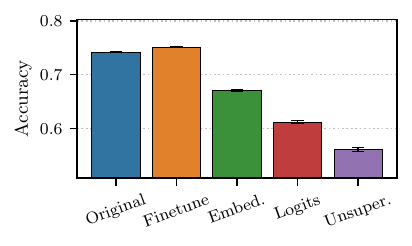}
        \caption{EMNIST}
        \label{fig:accuracy_comparison_emnist}
    \end{subfigure}

    \begin{subfigure}[b]{0.4\textwidth}
        \centering
        \includegraphics[width=\textwidth]{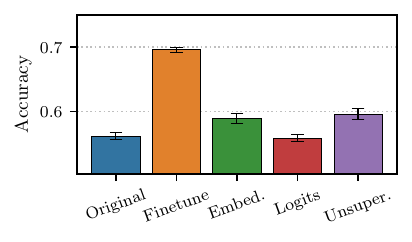}
        \caption{ModelNet10}
        \label{fig:accuracy_comparison_modelnet}
    \end{subfigure}
    \begin{subfigure}[b]{0.4\textwidth}
        \centering
        \includegraphics[width=\textwidth]{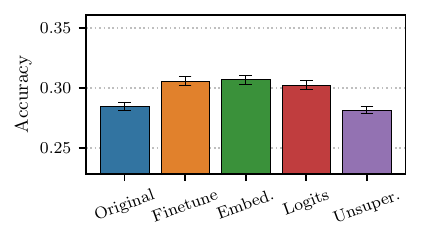}
        \caption{TU Berlin}
        \label{fig:accuracy_comparison_tuberlin}
    \end{subfigure}
    \caption[Comparison of supervised methods on different layers]{Accuracy comparison of supervised methods using inputs from different network layers, alongside baseline comparisons. Showing models on logits and embeddings performing worse than the finetuned model.}%
    \label{fig:supervised_layer_comparison_all}

\end{figure}

\paragraph{Details for learned energy applied to later layers}
To test whether information loss is a concern we train a classifier that predicts if the main model has the correct output for a given input on different layers of the base classifier.
We vary its input representation, considering logits (Logits), embeddings before the head extracted from the base model (Embeddings), and data directly (Original). 
We additionally include a model initialized from the base weights with a replaced head, operating on the data directly (Finetune). 
The results are shown in \Cref{fig:supervised_layer_comparison_all}.
On EMNIST, deeper layer features perform worse, suggesting information loss as a problem.
On ModelNet10 the network requires fine-tuning for good accuracy, indicating that the learned energy approach does not converge well when training from scratch. 
The fine-tuned model generally still outperforms embedding-based methods, indicating that the setup is inadequate. 
The exception is TU Berlin, where accuracy did not differ significantly between the pretrained model and using embeddings.
Since pretrained features generally perform worse in later layers, solely relying on them is likely suboptimal.

We use different architectures depending on the layer and type of input data.
The original architecture refers to using a modified version of the base architecture by replacing the head with a single-output linear layer while still training on data directly.
For the models trained on later layers on image data, the architecture consists of two 256-channel convolutional layers with GELU activations followed by adaptive pooling and a linear layer with hidden size 512 and GELU activation, followed by a final linear layer.
For the other models, we use 3 linear layers with 512 channels and GELU activations, followed by a final linear layer.
These are applied to either the logits or the embeddings of the final block or a previous layer in the MLP for LSTM (Layer 1 Inputs) and PointNet++ (Layer 4 Inputs).
The fine-tuned model copies the base model, replaces the head, and fine-tunes the head for 2 epochs, followed by fine-tuning the whole model for the remaining epochs.
Thus, a better training procedure for the learned energy model is required to achieve better results, for example by encouraging feature learning by using an auxiliary classification head and loss.

\subsection{Restricted Search Spaces}\label{app:restricted_search_space}

Applying optimization to continuous affine transformations is challenging because the search space is non-compact.
For instance, the matrix representation of scale transformations, with the topology inherited from $\mathbb{R}^n$, is unbounded and thus non-compact according to the Heine-Borel theorem \citep[Theorem 2.41]{Rudin1976}.
This poses a challenge for search algorithms, as they may need to explore an unbounded space.
In the literature, we identified two solutions.

The first solution assumes a decomposition of the affine group into subgroups. The unbounded subgroups, such as scaling and shearing, are restricted to a specific domain. This approach is taken by ITS\citep{ITS}.
While this ensures the feasibility of the search, a limitation is the assumption that the global minimum resides within the restricted domain.
As this may not always hold, it deviates from strict canonicalization since some transformations might remain uncorrected.
It is a heuristic approach that only finds approximate solutions in a limited domain.
However, it may be advantageous if it helps avoid regions of local minima that exhibit low energy but are misclassified by the classifier.

\begin{figure}[t]
    \centering
    \begin{subfigure}{0.24\textwidth}
        \centering
        \includegraphics[width=\linewidth]{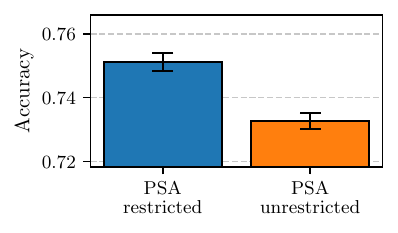}
        \caption{Energy (60)}
        \label{fig:exp1_1_energy}
    \end{subfigure}
    \hfill
    \begin{subfigure}{0.24\textwidth}
        \centering
        \includegraphics[width=\linewidth]{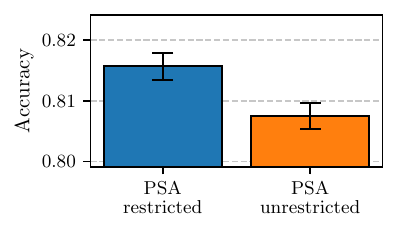}
        \caption{PC-kNN (60)}
        \label{fig:exp1_1_knn}
    \end{subfigure}
    \hfill
    \begin{subfigure}{0.24\textwidth}
        \centering
        \includegraphics[width=\linewidth]{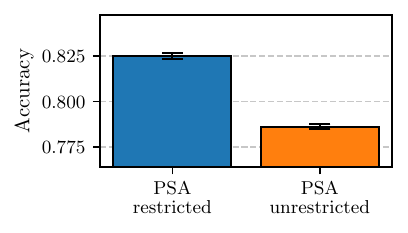}
        \caption{Energy (480)}
        \label{fig:exp1_1_energy_more_budget}
    \end{subfigure}
    \hfill
    \begin{subfigure}{0.24\textwidth}
        \centering
        \includegraphics[width=\linewidth]{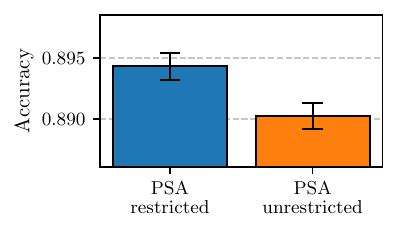}
        \caption{PC-kNN (480)}
        \label{fig:exp1_1_knn_more_budget}
    \end{subfigure}
    \caption{Accuracy after canonicalization with Simulated Annealing on MNIST, with and without domain restriction, at 60 and 480 total evaluations.}
    \label{fig:exp1_1}
\end{figure}

The second approach requires a topologically coercive energy function\citep{Shumaylov2025}.
Formally, given an energy function $E: X \to [0, \infty)$, this means that for every $r \in \mathbb{R}_{+}$ there exists a compact set $K_r \subseteq X$ such that $E(x) > r$ for all $x \in X \setminus K_r$.
This approach does not assume a restricted domain, but rather that the solution resides within a searchable compact set, which might not hold for all scores.
For image data, extreme scaling and shearing typically result in information being moved outside the grid, which should be detected as OOD.
Non-closedness is less critical, as one can utilize the closure (including limit points) \citep{Shumaylov2025} as a solution.
We empirically evaluate whether restricting the search space improves performance using SA on padded MNIST, with and without projecting back into the domain.
Performance is expected to degrade significantly when scores lack topological coercivity, as the search may extend far beyond the domain of the transformations of the data.
Even with topological coercivity, domain restriction may be beneficial by focusing samples in regions where the minimum is likely located.
\Cref{fig:exp1_1} shows the results when using 10 parallel runs with 12 evaluations per run.
The step count is low, but in our experiments this performed better than using more steps with fewer parallel runs.
The neighborhood size was set to 0.25 of the parameter domain, or 0.25 for unbounded parameters.
Domain restriction generally improves performance for both energy-based and PC-kNN scores (3-nearest-neighbor, cosine distance on final-layer features).

\begin{figure}[t]
    \centering
    \includegraphics[width=0.7\textwidth]{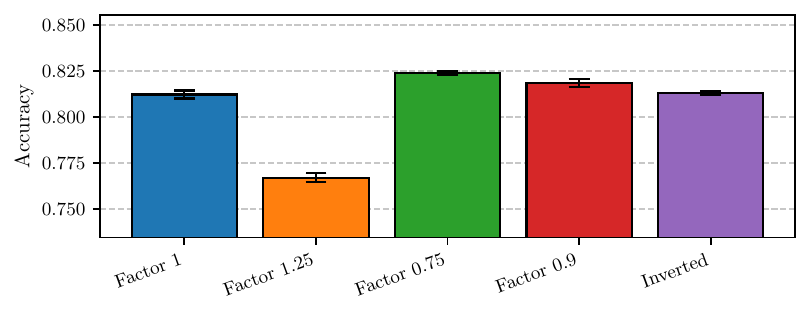}
    \caption{Comparison of different domain sizes and inverted domain using energy of the logits and random search (budget 60). Smaller domains can be beneficial.}
    \label{fig:resnet_small_random_energy}
\end{figure}

We repeated the experiment with 20 runs and 24 steps each to reduce the influence of the lower sample density caused by the increased domain size when not restricting the domain.
Here, the kNN-based detector still performs slightly worse when not restricting the domain, but the gap is smaller.
The energy-based detector still benefits significantly from domain restriction, suggesting it likely lacks topological coercivity.
As domain restriction consistently improved performance, it should be treated as a hyperparameter when using unsupervised OOD scores for canonicalization.
The search domain must cover the data's transformation range to allow exact inversion, but an overly large domain wastes samples and admits spurious minima. 
Since the true range is usually unknown, bounds must be set manually. 
We use the same domain as the data augmentation but do not invert the final matrix, which stays close to the actual distribution under the chosen decomposition.
This is not necessarily the optimal domain.
We consider domains where we reduce the maximum shear and scale by a specific factor(\Cref{fig:resnet_small_random_energy} for MNIST).
The transformations that would be able to perfectly undo the transformation (inverted) are outperformed by transformations that do not invert the final matrix and use smaller scaling and shearing.

\subsection{Budget Comparisons}
\label{app:budget}

\begin{figure}[h!]
    \centering
    \begin{subfigure}{\textwidth}
        \centering
        \raisebox{0.5\height}{\rotatebox{90}{\small\textbf{(a) MNIST}}}\hspace{2pt}%
        \includegraphics[width=0.8\textwidth]{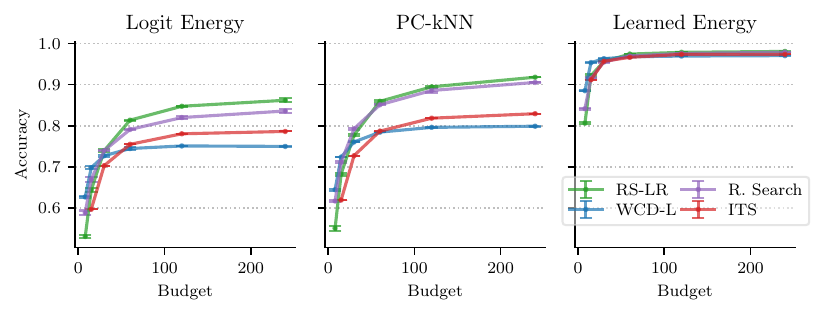}
        \phantomcaption
        \label{fig:budget_mnist}
    \end{subfigure}

    \begin{subfigure}{\textwidth}
        \centering
        \raisebox{0.2\height}{\rotatebox{90}{\small\textbf{(b) ModelNet10}}}\hspace{2pt}%
        \includegraphics[width=0.8\textwidth]{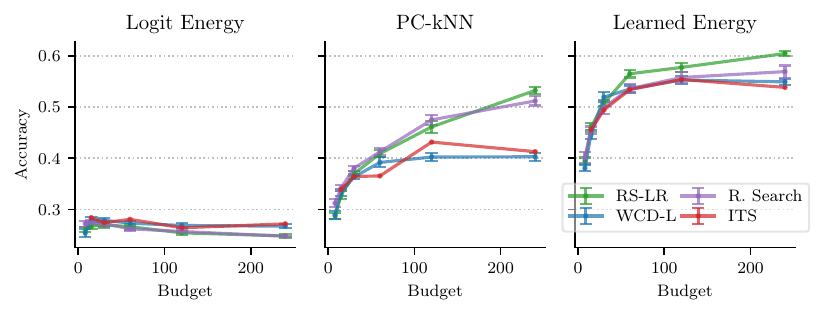}
        \phantomcaption
        \label{fig:budget_modelnet}
    \end{subfigure}

    \begin{subfigure}{\textwidth}
        \centering
        \raisebox{0.3\height}{\rotatebox{90}{\small\textbf{(c) TU Berlin}}}\hspace{2pt}%
        \includegraphics[width=0.8\textwidth]{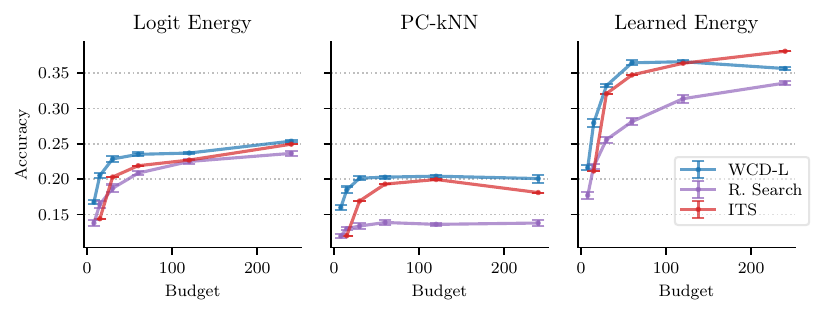}
        \phantomcaption
        \label{fig:budget_tuberlin}
    \end{subfigure}

    \caption[Accuracy comparison when scaling the search budget]{Accuracy after canonicalization when scaling the budget of different optimization algorithms on multiple datasets and scores.}
    \label{fig:exp2_2_accuracy}
\end{figure}

Additional scaling results for MNIST, TU Berlin, and ModelNet10 are shown in \Cref{fig:exp2_2_accuracy}.
TU Berlin is an exception where coordinate descent-based algorithms, and especially ITS, outperform global search algorithms even at larger budgets.
This suggests that the bias introduced by the coordinate ordering in ITS can assist in finding parameters leading to correctly classified points.
On TU Berlin, the OOD scores performed worst, resulting in lower overall accuracy.
The other datasets show similar trends to those in the main part, see \Cref{sec:exp:search}: random search is best at larger budgets, local refinement gives slight improvements (notably on ModelNet10), and for lower budgets, weighted coordinate descent outperforms random search.

\subsection{Energy Landscapes}
\label{app:energy_landscapes}

\begin{figure}[t]
    \centering
    \begin{subfigure}{0.32\linewidth}
        \centering
        \includegraphics[width=\linewidth]{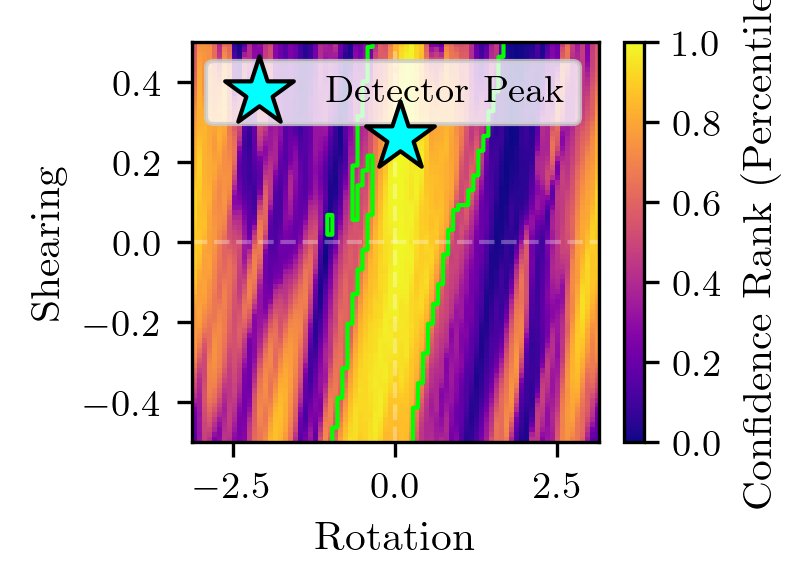}
        \caption{kNN}
    \end{subfigure}
    \hfill
    \begin{subfigure}{0.32\linewidth}
        \centering
        \includegraphics[width=\linewidth]{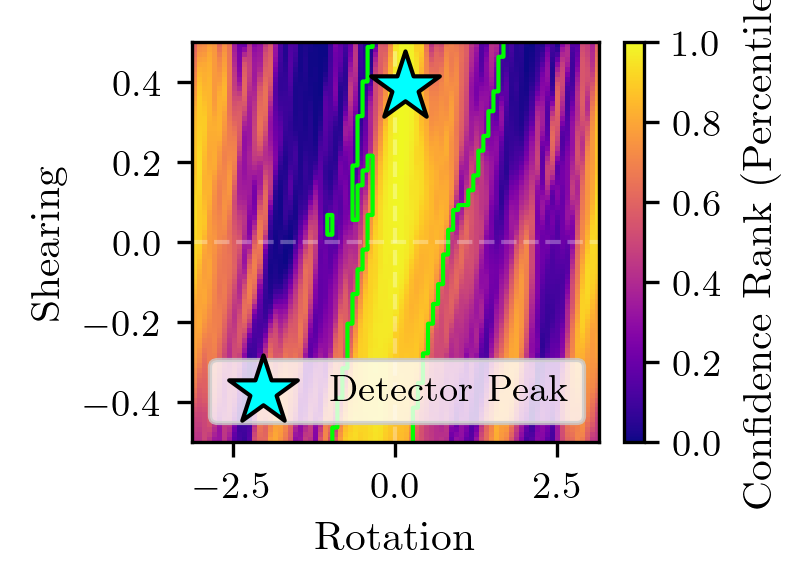}
        \caption{Energy}
    \end{subfigure}
    \hfill
    \begin{subfigure}{0.32\linewidth}
        \centering
        \includegraphics[width=\linewidth]{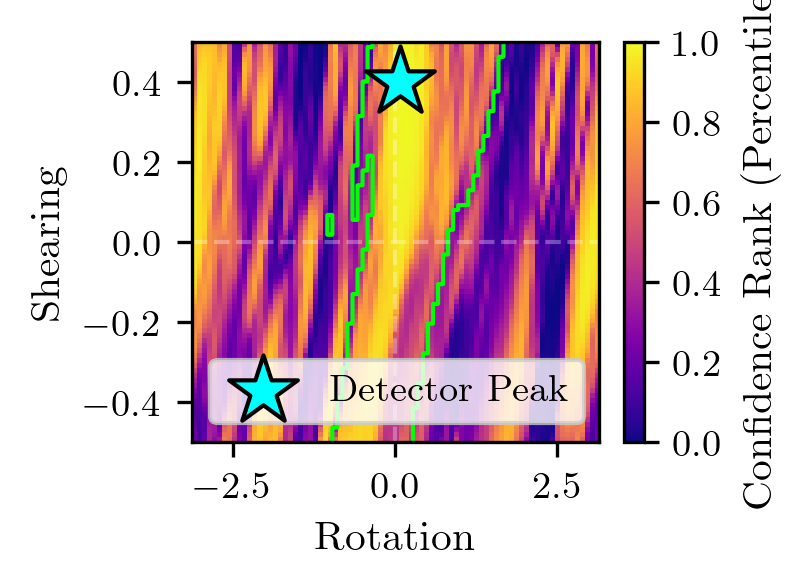}
        \caption{Entropy}
    \end{subfigure}

    \vspace{0.6em}

    \begin{subfigure}{0.32\linewidth}
        \centering
        \includegraphics[width=\linewidth]{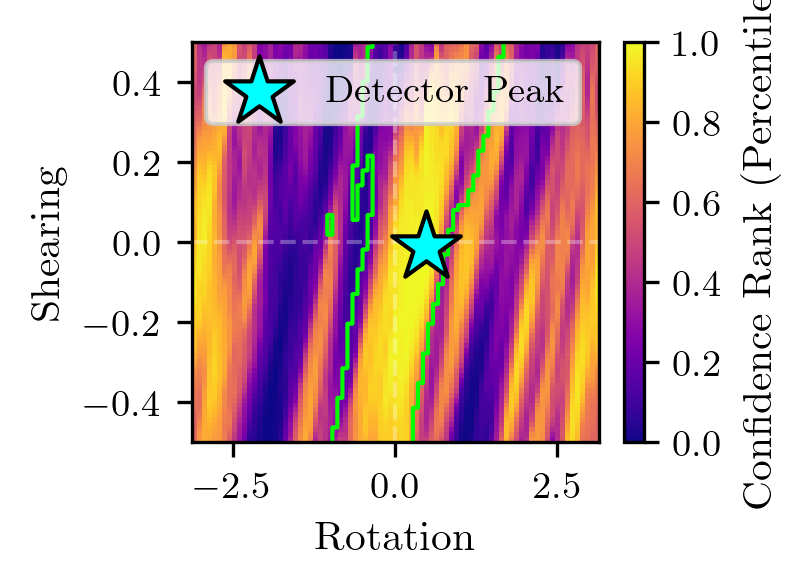}
        \caption{Mahalanobis}
    \end{subfigure}
    \hfill
    \begin{subfigure}{0.32\linewidth}
        \centering
        \includegraphics[width=\linewidth]{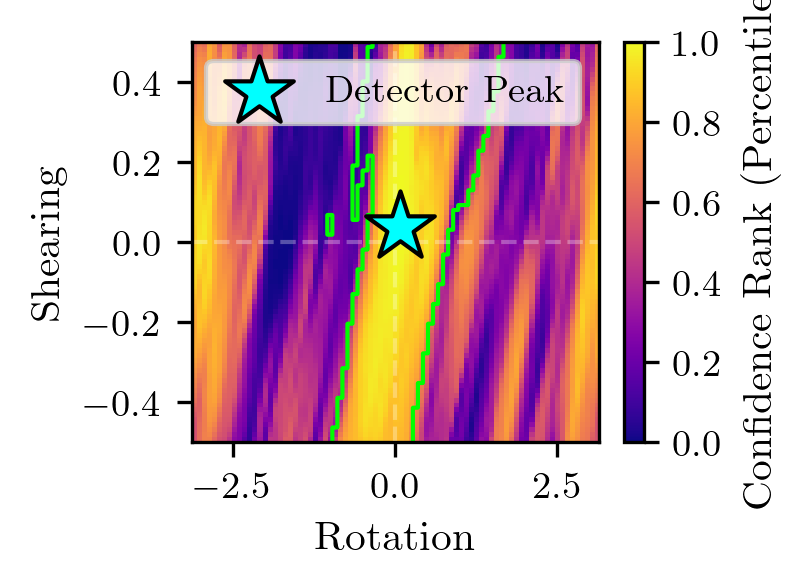}
        \caption{ViM}
    \end{subfigure}
    \hfill
    \begin{subfigure}{0.32\linewidth}
        \centering
        \includegraphics[width=\linewidth]{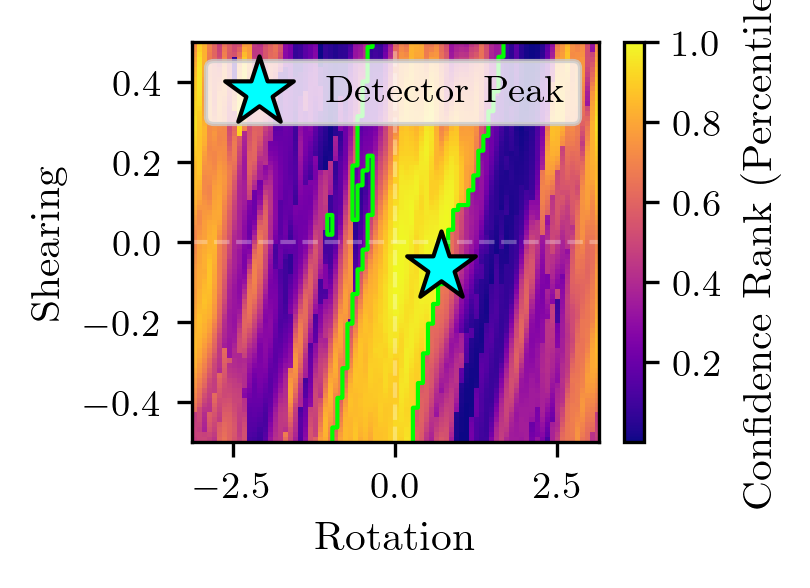}
        \caption{PC-kNN}
    \end{subfigure}

    \caption{Score landscapes of different OOD scores on an EMNIST sample, varying rotation and the first shear factor (other transformations fixed). Values are percentile ranks. The correctly classified region is overlaid.}
    \label{fig:bigger_emnist_landscapes}
\end{figure}

We selected a random untransformed sample from EMNIST and varied rotation and the first shearing factor, keeping the other transformations fixed. We normalize the scores by calculating the ranking of the scores and displaying the percentile of the scores. This is because random search relies on score comparisons, making relative ranking more relevant.
This is shown in \Cref{fig:bigger_emnist_landscapes} for some scores, where we also include the region where the samples are correctly classified. It shows that for this sample, the energy landscapes have similar regions where the scores are high. However, the region where the minima are located can vary much more. Per-class variants of kNN are discontinuous, as a change in the predicted class can abruptly alter the distance.
Even with 3 of the 5 dimensions fixed, the landscapes show many local minima, necessitating global optimization methods. All algorithms show similarities in the regions where they assign higher scores.
Theoretically it suffices for the minima to lie in a correctly classified region, but on a limited budget a narrow low-value region may be missed. 
A broad low-value region overlapping the correctly classified area, with no competing low minima elsewhere, is therefore preferable for optimization.

\subsection{Selection Gate Thresholds}\label{app:selectiongate}

\begin{figure}[t]
  \centering
  \begin{subfigure}[b]{0.48\textwidth}
    \centering
    \includegraphics[width=\linewidth]{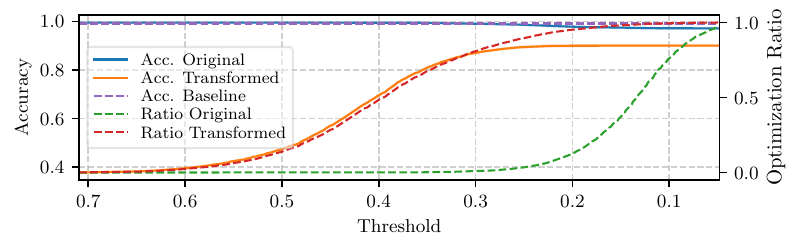}
    \caption{MNIST (unsupervised)}
    \label{fig:tradeoff_mnist}
  \end{subfigure}
  \hfill
  \begin{subfigure}[b]{0.48\textwidth}
    \centering
    \includegraphics[width=\linewidth]{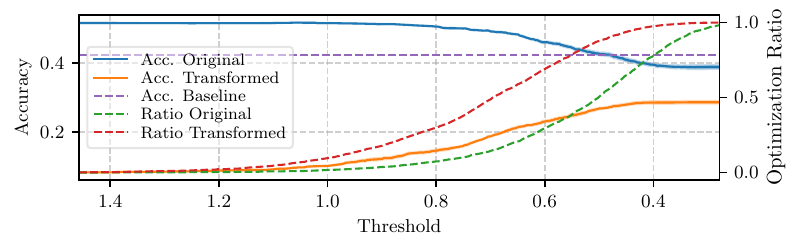}
    \caption{TU Berlin (unsupervised)}
    \label{fig:tradeoff_tuberlin}
  \end{subfigure}

  \vspace{0.6em}

  \begin{subfigure}[b]{0.48\textwidth}
    \centering
    \includegraphics[width=\linewidth]{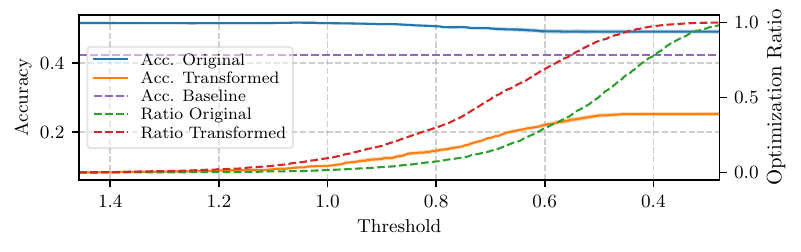}
    \caption{TU Berlin (margin 0.1)}
    \label{fig:tradeoff_tuberlin_margin}
  \end{subfigure}
  \hfill
  \begin{subfigure}[b]{0.48\textwidth}
    \centering
    \includegraphics[width=\linewidth]{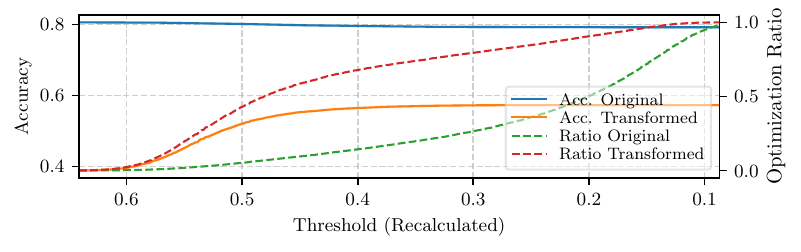}
    \caption{SI-Score rotation (margin 0.1)}
    \label{fig:tradeoff_SI-Score}
  \end{subfigure}

  \vspace{0.6em}

  \begin{subfigure}[b]{0.48\textwidth}
    \centering
    \includegraphics[width=\linewidth]{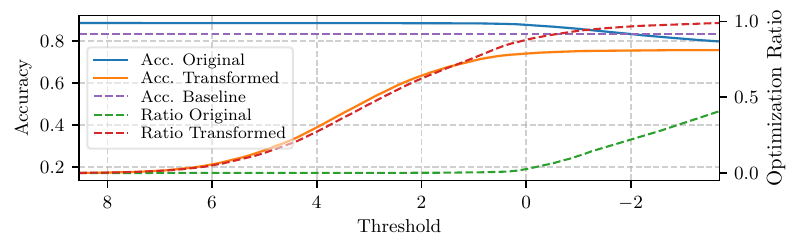}
    \caption{EMNIST (learned energy)}
    \label{fig:tradeoff_emnist_supervised}
  \end{subfigure}
  \hfill
  \begin{subfigure}[b]{0.48\textwidth}
    \centering
    \includegraphics[width=\linewidth]{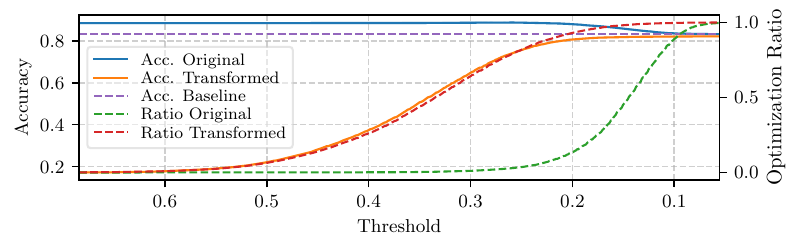}
    \caption{EMNIST (augmented model)}
    \label{fig:tradeoff_emnist_augmented}
  \end{subfigure}

  \caption{
  Trade-off between accuracy on the original and transformed test sets as the OOD threshold is varied.
  (a, b) Plain unsupervised gating with the best detector per dataset. 
  (c, d) Accepting canonicalization only when it lowers the normalized score by a margin of 0.1 improves the trade-off.
  (e, f) On EMNIST, switching between the vanilla and augmented classifier with an unsupervised detector.}
  \label{fig:tradeoff_all}
\end{figure}

\Cref{fig:tradeoff_all} shows the trade-off for the remaining datasets, using the best unsupervised detector from \Cref{fig:comparision_detectors_1}. The baseline is the augmented model on original data.
MNIST allows higher accuracy on transformed data with only minor loss on untransformed data, while TU Berlin shows a more symmetric trade-off.
In the main part we accepted any canonicalized sample whose OOD score beat the transformed input. The trade-off improves if we instead require a margin: normalizing scores by the validation-set range and accepting only when the canonicalized score is 0.1 lower improves TU Berlin's trade-off, though peak accuracy on transformed data drops.
This margin is also needed on more complex datasets. For SI-Score (rotation) with a ViT-B16, accepting at a 0.1 margin gives a workable trade-off, but since background changes make SI-Score easy to flag as OOD, the trade-off may be worse on more realistic data. Original accuracy refers to the ImageNet validation accuracy, as test labels are unavailable.
On EMNIST, switching between the vanilla and augmented classifier using an unsupervised detector outperforms learned-energy canonicalization with thresholding, so a secondary model trained on transformed data is generally more effective than canonicalization.

\end{document}